\documentclass{article}

\usepackage{arxiv}

\usepackage[utf8]{inputenc} 
\usepackage[T1]{fontenc}    
\usepackage{hyperref}       
\usepackage{url}            
\usepackage{booktabs}       
\usepackage{amsfonts}       
\usepackage{nicefrac}       
\usepackage{microtype}      
\usepackage{lipsum}		

\usepackage{doi}
\usepackage{subfigure}
\usepackage[dvipsnames]{xcolor}
\usepackage{amsmath}
\usepackage{tabularx}
\usepackage{multirow}
\usepackage{graphicx}
\usepackage{enumerate}
\usepackage{natbib}
\usepackage{amssymb}
\usepackage[figuresright]{rotating}
\usepackage{amsthm}
\usepackage{setspace}

\usepackage[ruled,vlined]{algorithm2e}
\usepackage{comment}

\title{Augmented Learning of Heterogeneous Treatment Effects via Gradient Boosting Trees}

\author{ \href{https://orcid.org/0000-0002-2093-8261}{\includegraphics[scale=0.06]{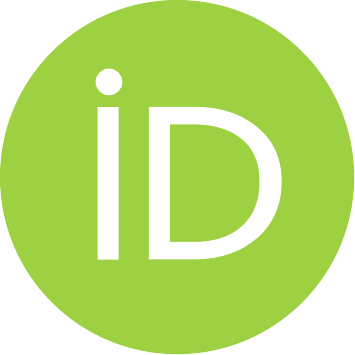}\hspace{1mm}Heng Chen} \\
	Department of Biostatistics\\
	Gilead Sciences, Inc.\\
	Foster City, CA 94404 \\
	\texttt{Henry.Chen10@gilead.com} \\
	\And
 James Y. Dai\thanks{Corresponding author} \\
	Public Health Sciences Division/Department of Biostatistics\\
	Fred Hutchinson Cancer Center/University of Washington,
	Seattle, WA\\
	\texttt{jdai@fredhutch.org}\\
 \And
  Michael L. LeBlanc\\
	Public Health Sciences Division/Department of Biostatistics\\
	Fred Hutchinson Cancer Center/University of Washington,
	Seattle, WA\\
	\texttt{mleblanc@fredhutch.org}\\
}

\renewcommand{\shorttitle}{Augmented Learning of HTE}

\hypersetup{
pdftitle={Augmented Learning of HTE},
pdfsubject={},
pdfauthor={Heng Chen, James Y. Dai, Michael L. LeBlanc},
pdfkeywords={Augmentation, Functional Optimization,  Marker-treatment Interaction, Permutation Test , Robustness and Efficiency.},
}

\begin{document}
\maketitle

\begin{abstract}
	Heterogeneous treatment effects (HTE) based on patients' genetic or clinical factors are of significant interest to precision medicine. Simultaneously modeling HTE and corresponding main effects for randomized clinical trials with high-dimensional predictive markers is challenging. Motivated by the modified covariates approach, we propose a two-stage statistical learning procedure for estimating HTE with optimal efficiency augmentation, generalizing to arbitrary interaction model and exploiting powerful extreme gradient boosting trees (XGBoost). Target estimands for HTE are defined in the scale of mean difference for quantitative outcomes, or risk ratio for binary outcomes, which are the minimizers of specialized loss functions. The first stage is to estimate the main-effect equivalency of the baseline markers on the outcome, which is then used as an augmentation term in the second stage estimation for HTE. The proposed two-stage procedure is robust to model mis-specification of main effects and improves efficiency for estimating HTE through nonparametric function estimation, e.g., XGBoost. A permutation test is proposed for global assessment of evidence for HTE. An analysis of a genetic study in Prostate Cancer Prevention Trial led by the SWOG Cancer Research Network, is conducted to showcase the properties and the utilities of the two-stage method.
\end{abstract}


\section{Introduction}

\label{sec:intro}

Precision medicine is an emerging area of biomedical research aiming to tailor therapies or prevention toward patients' genetic or clinical backgrounds, as opposed to the conventional ``one size fits all'' approach. Pivotal to implementing precision medicine is to estimate individual or subgroup treatment effects that are modified by baseline ``predictive markers'', i.e., the markers that predict heterogeneous treatment effects (HTE) among trial participants. \citep{evans2003pharmacogenomics,weinshilboum2004pharmacogenomics}\citep{ simon2013implementing} Identification of these predictive markers can also enhance the understanding of the mechanisms of treatment effects on clinical end points.

When there is a large number of candidate predictive markers, it is challenging to efficiently and robustly estimate heterogeneous treatment effects in a randomized clinical trial (RCT). Sample size of a RCT is driven by the primary intent-to-treat analysis of the overall treatment effect, therefore rarely adequate for assessing high-dimensional predictive markers. Moreover, simultaneously modeling high-dimensional interactions and main effects is challenging. Let $Y$ denote the trial end point, $T$ denote the randomized treatment assignment (typically 1 for treatment and 0 for control), and $X$ denote a vector of predictive markers (with 1 as the first element for the intercept). A common parametric working model for $(Y,T,X)$ is
\begin{equation}
    g[\mathbb{E}(Y|X,T)]=\mathcal{A}(X;\boldsymbol{\alpha})+\mathcal{F}(X;\boldsymbol{\beta})T,
    \label{generalmodel}
\end{equation}
where $g$ is a link function, $\mathcal{A}$ and $\mathcal{F}$ represent functions of the main effect and the HTE with respective parameter $\boldsymbol{\alpha}$ and $\boldsymbol{\beta}$. More general models may include classes of nonparametric functional modeling of $\mathcal{A}$ and $\mathcal{F}$. While the HTE component $\mathcal{F}$ is of primary interest, its estimation is dependent on properly accounting for the main effect component $\mathcal{A}$. Specifically, $\mathcal{A}$ sets the baseline of $Y$ upon which the HTE is defined, and modeling $\mathcal{A}$ improves efficiency for estimating $\mathcal{F}$, yet mis-specification of $\mathcal{A}$ may draw bias to the estimator of $\mathcal{F}$. \citep{vanderweele2012sensitivity} To this end, statistical convention is to impose constraints between $\mathcal{A}(X;\boldsymbol{\alpha})$ and $\mathcal{F}(X;\boldsymbol{\beta})$: an interaction term is present in the model only if the corresponding main effect is also included, so-called “hierarchy” or “heredity” property of interaction and main effect. \citep{cox1984interaction,peixoto1987hierarchical,choi2010variable} When $X$ is high-dimensional, model selection and adaptive estimation of $\mathcal{F}(X;\boldsymbol{\beta})$ become challenging, because one has to simultaneously estimate $\mathcal{A}(X;\boldsymbol{\alpha})$ under the hierarchy constraint. See, for example, extensions of LASSO to estimate interactions in parametric models. \citep{bien2013lasso,lim2015learning} Nonparametric methods have been proposed for estimating HTE, for instances recursive partitioning via trees and random forests, \citep{su2009subgroup,athey2016recursive,wager2018estimation} essentially forcing the same tree structures for $\mathcal{A}$ and $\mathcal{F}$. Alternatively, the case-only random forest method has been proposed to estimate HTE. \citep{dai2019case,dai2019caseonly}

Assuming an equal randomization probability to the treatment arm or the control arm, the modified covariates approach proposed by \cite{tian2014simple} exploits the randomization and removes the main-effect component $\mathcal{A}$. The heterogeneous treatment effects are modeled by linear additive terms $XT$ as  $g[\mathbb{E}(Y|X,T)]=\boldsymbol{\beta}^T (XT)$, where $T$ is coded as $+1$ or $-1$ for treatment or control with 0.5 probability, and $XT$ is the modified covariates. One caveat is that, except for the identity function, the modified covariates model is generally misspecified and does not necessarily result in the HTE in the scale of the link function $g$. For high-dimensional $X$, \cite{tian2014simple} used the regularized model selection for the HTE component $\mathcal{F}$ via LASSO. This approach also enables other nonparametric learning methods such as extreme gradient boosting trees (XGBoost) for estimating individualized treatment effects. \citep{sugasawa2019estimating} To gain efficiency, \cite{tian2014simple} proposed an augmentation term to the estimating function, in the same spirit of semiparametric models that improve efficiency by including auxiliary covariates. \citep{tsiatis2007semiparametric,tsiatis2008covariate,zhang2008improving} A special case is simple linear models, where the augmented term is equivalent to the main-effect component, essentially connecting back to the full model (\ref{generalmodel}). \cite{chen2017general} generalized the modified covariates approach to a general statistical framework for subgroups identification via weighting and A-learning with various implementations in \cite{huling2018subgroup}. 

Motivated by the modified covariates approach and the aforementioned semiparametric theory, we propose a two-stage statistical learning procedure that permits adaptive nonparametric modeling of both mean-difference and risk-ratio estimands for HTE. In contrast to the standard maximum likelihood estimation for the joint model (\ref{generalmodel}), we define the mean-difference and risk-ratio estimands for HTE by minimizing specialized loss functions of the modified covariates, and the main effects for predictive variables are treated as a robust augmentation term for improving the efficiency of estimating HTE. Misspecified augmentation does not affect consistent estimation of HTE, therefore eliminating the dependency of HTE on the main-effect modeling. Three additional novel contributions have been made: first, optimal augmentation terms for both mean-difference and risk-ratio HTE can be derived for an arbitrary interaction model, where both \cite{tian2014simple} and \cite{chen2017general} need to assume zero interaction effect to derive the optimal augmentation term; second, a novel permutation test is devised to assess the significance of HTE conditional on the estimated main effect component; third, a weighting method is applied to estimate the
optimal augmentation term,  circumventing the need of estimating two separate conditional means as implemented in \cite{tian2014simple}, \cite{chen2017general} and \cite{huling2018subgroup}.

One more benefit of operationally separating HTE and main-effect augmentation is that, for high-dimensional covariates, we are capable of exploiting powerful nonparametric learning algorithms, e.g., eXtreme GBT (XGBoost) for adaptive learning both main effect and HTE, which has not been attempted previously. \cite{friedman2001greedy} proposed GBT through steepest descent optimization in functional space. \cite{buhlmann2003boosting} and \cite{buehlmann2006boosting} investigated L-2 boosting algorithms for high-dimensional predictors and evaluated the consistency property. \cite{chen2016xgboost} proposed eXtreme GBT (\texttt{XGBoost}) with regularization to penalize for complex tree structures, reducing the computational time for boosting trees in any pre-specified convex loss function. In this work we propose a novel two-stage gradient boosting trees algorithm via \texttt{XGBoost} for sequentially learning of main effect and HTE.

This article is organized as follows. Section \ref{method} introduces HTE estimands and the modified covariates approach. Section \ref{ourmethod} describes the proposed augmented learning procedure for estimating heterogeneous treatment effects, separately for mean-difference and risk-ratio estimands of HTE. Section \ref{permutation-metho} presents a permutation test for heterogeneous treatment effects. Section \ref{tsgbt-metho} describes the implementation of the procedure by adapting \texttt{XGBoost}.  Section \ref{sec:simulations} presents extensive simulation studies for comparing the efficiency of our approach with existing methods, and assessing the validity of the permutation test. 
Section \ref{sec:PCCT} showcases the proposed method in a genetic study for Prostate Cancer Prevention Trial (PCPT) trial. Section \ref{sec:disc} concludes with a discussion.

\vspace{-0.7cm}
\section{Methods}
\vspace{-0.2cm}
\label{sec:methodology}
\subsection{HTE estimands}
\label{method}
Let $Y$ denote a continuous or binary response, $T$ denote a binary indicator for treatment assignment, coded by $-$1 for the control arm and +1 for the experiment arm. Let $X$ represent a $p$-dimensional vector of baseline covariates that may modify treatment effect. The treatment is randomly assigned to study participants with $0<\text{Pr}(T=1)<1$, so that $T$ is independent of $X$. Following the potential outcomes framework of \cite{neyman1923applications} and \cite{rubin1974estimating}, we denote $Y^{(-1)}$ and $Y^{(1)}$ to be the potential response had a participant been assigned to the control or the experiment, respectively. We can only observe one of the paired potential outcomes, $Y^{(t)}$, which is the observed response $Y$ given the realized treatment assignment $T=t$. Assume the observed data consist of independently and identically distributed random vectors $(Y_i,T_i,X_i), \text{\ for\ } i=1,\dots,n$ participants.

The estimand for the heterogeneous treatment effect (HTE) is defined as a function of $X=x$. For a continuous response or a dichotomized outcome, we define HTE as
\begin{equation}
    \tau(x)=\mathbb{E}(Y^{(1)}-Y^{(-1)}|X=x)= \mathbb{E}(Y|X=x,T=1) -  \mathbb{E}(Y|X=x,T=-1),   \label{estimand1}
\end{equation}
the difference of the expectations of $Y$ given $X$ in either treatment arm. For a binary outcome, a risk ratio estimand for HTE can be defined as
\begin{equation}
    \tau(x)=\frac{\mathbb{E}(Y^{(1)}|X=x)}{\mathbb{E}(Y^{(-1)}|X=x)} = \frac{ \mathbb{E}(Y|X=x,T=1)}{\mathbb{E}(Y|X=x,T=-1) } . \label{estimand2}
\end{equation}
Of primary interest in this article is the estimation of and the inference for the function $\tau(x)$. If $\tau(x)$ reduces to a constant, there is no heterogeneity and every participant has the same treatment effect --- the null hypothesis for testing HTE.

The standard maximum likelihood approach assumes a probabilistic model as in (\ref{generalmodel}). When $X$ is moderate- or high-dimensional, it is challenging to simultaneously model both main-effect component $\mathcal{A}(X)$  and the HTE function $\mathcal{F}(X)$ by flexible nonparametric functions that conform to the heredity constraint. Instead, we define loss functions that directly target at the HTE estimand (\ref{estimand1}) and (\ref{estimand2}) by functional estimation, while the main-effect component is treated separately as an augmentation term in the loss function. This two-stage statistical learning procedure is motivated by the modified covariates approach \citep{tian2014simple}, where $T$ is coded to be $+1$ and $-1$ with an equal probability. A succinct summary of the modified covariates approach is presented in Supplement S1. 




\subsection{Two-stage nonparametric estimation of HTE}
\label{ourmethod}
\subsubsection{Mean-difference estimand for a continuous response}
\label{continuousresponse}
Suppose the randomization probability for $T=+1$ or $T= -1$ may not equal to $\frac{1}{2}$.  To obtain the mean-difference estimand of HTE as defined in (\ref{estimand1}), we minimize the following expected squared-error loss function that is weighted by the inverse of the randomization probabilities,
\begin{equation}
  \underset{\mathcal{F}}{\text{argmin}} \mbox{\hspace{10pt}} \mathbb{E} \Bigg[\Big\{\frac{T+1}{2\text{Pr}(T=1)}-\frac{T-1}{2\text{Pr}(T=-1)}\Big\}  \Big\{Y- \mathcal{F}(X) T\Big\}^2 \Bigg].
   \label{weighted_conti_loss}
\end{equation}
Note that $\mathcal{F}(X)$ is now defined to be a general function without parameter $\boldsymbol{\beta}$.  By this inverse probability weighted construct, we define the HTE estimand in a ``hypothetical population'' in which treatment labels are equally assigned. This facilitates the derivation of the estimand and the optimal augmentation. The estimating equation for the minimizer of  (\ref{weighted_conti_loss}) is
\begin{equation}
    \mathbb{E}\Big\{S(Y, \mathcal{F}(X)T) \Big\}=0,
    \label{weighted_conti_score}
\end{equation}
where
$S(Y, \mathcal{F}(X)T) = \bigg\{\frac{T+1}{2\text{Pr}(T=1)}+\frac{T-1}{2\text{Pr}(T=-1)}\bigg\} \bigg\{Y- \mathcal{F}(X) T\bigg\}$
is the first-order derivative of the loss function (\ref{weighted_conti_loss}) with respect to $\mathcal{F}(X)$, ignoring the constant factor $-2$, canceling out the factor $T$ by flipping the negative sign in $\Big\{\frac{T+1}{2\text{Pr}(T=1)}-\frac{T-1}{2\text{Pr}(T=-1)}\Big\}$ and assuming the exchangeability of derivative and expectation.  Integrating  estimating function (\ref{weighted_conti_score}) over the conditional distribution of $T$ conditional on $X$, we obtain
$\mathbb{E}\left(Y|X,T=1\right) - \mathcal{F}(X) - \mathbb{E}\left(Y|X,T=-1\right) - \mathcal{F}(X) = 0.$
Then the minimizer is one half the mean-difference estimand,
\begin{equation*}
   \mathcal{F}^{*}(x)=\frac{1}{2}\Big\{\mathbb{E} (Y|X=x,T=1) - \mathbb{E} (Y|X=x,T=-1)\Big\}.
\end{equation*}

To improve efficiency for estimating the mean-difference estimand, an inverse probability weighted augmentation term can be added to (\ref{weighted_conti_score}), namely
$a(X)\bigg\{\frac{T+1}{2\text{Pr}(T=1)}+\frac{T-1}{2\text{Pr}(T=-1)}\bigg\}$,
whose expectation is zero. The augmented estimating equation is,
\begin{equation}
    \mathbb{E}\Big\{S^{\text{aug}}(Y, \mathcal{F}(X)T)\Big\}=0,
    \label{aug_weighted_conti_est_eq}
\end{equation}
where
$S^{\text{aug}}(Y, \mathcal{F}(X)T) = S(Y, \mathcal{F}(X)T) - a(X)\bigg\{\frac{T+1}{2\text{Pr}(T=1)}+\frac{T-1}{2\text{Pr}(T=-1)}\bigg\}$.
The solutions of (\ref{weighted_conti_score}) and (\ref{aug_weighted_conti_est_eq}) are the same because the expectation of the augmentation is zero. To obtain the optimal augmentation term $a_0(X)$ that minimizes the variance of the estimator for $\mathcal{F}(X)$, it is asymptotically equivalent to minimize the variance of the augmented estimating function $S^{\text{aug}}(Y, \mathcal{F}(X)T)$ \citep{robins1994estimation}. Derived in Appendix, the optimal augmentation term is $a_0(X)= \frac{1}{2}\Big\{\mathbb{E}(Y|X, T=1)+\mathbb{E}(Y|X, T=-1)\Big\}$ under any arbitrary interaction model, any randomization ratio and nonparametric learning of HTE. A special case is $\text{Pr}(T=1)=\frac{1}{2}$, where $a_0(X) = \mathbb{E}(Y|X)$. To avoid constructing two separate models for estimating $\mathbb{E}(Y|X, T=1)$ and $\mathbb{E}(Y|X, T=-1)$ under unequal randomization ratio, we can estimate $\mathbb{E}(Y|X, T=1)+\mathbb{E}(Y|X, T=-1)$ as one whole term via minimizing the below weighted loss function,
\begin{equation}
  \underset{\mathcal{A}}{\text{argmin}} \mbox{\hspace{10pt}} \mathbb{E} \Bigg[\Big\{\frac{T+1}{2\text{Pr}(T=1)}-\frac{T-1}{2\text{Pr}(T=-1)}\Big\}  \Big\{Y- \mathcal{A}(X)\Big\}^2 \Bigg],
   \label{weighted_aug_mse_loss}
\end{equation}
which can be implemented through any parametric, nonparametric function, or an ensemble learner.

\noindent
{\bf Remark.} {\it We make important remarks about the properties of the HTE estimator that solves the augmented estimating function (\ref{aug_weighted_conti_est_eq}). In essence, this approach defines HTE as the targeted estimand by solving a loss function and then adds the main effect back as the augmentation in estimating functions, eliminating the dependency between main effects and interactions. The special structure of $-1$ and $+1$ in two randomized treatment groups dictates the orthogonality between $T$ and $X$, because of which, the estimator for HTE is consistent regardless whether the estimation of the augmentation term is correctly specified or not. Moreover, valid inference of HTE can be drawn solely from the augmented estimating function (\ref{aug_weighted_conti_est_eq}), treating the plug-in estimator for the augmentation as a known quantity. These properties are in stark contrast to caveats of the standard interaction estimation \citep{vanderweele2012sensitivity}.
}

The derivation above also implies a two-stage procedure for solving the augmented estimation function: first, one can use any regression models under the weighted loss function (\ref{weighted_aug_mse_loss}) to estimate $a_0(X)$. A simple linear model may not be optimal but may well be adequate for the purpose of efficiency augmentation. In the second stage, an estimator $\widehat{\mathcal{F}}(X)$ can be obtained by minimizing
\begin{equation}
\underset{\mathcal{F}}{\text{argmin}}  \mbox{\hspace{10pt}} \frac{1}{n}\sum_{i=1}^n  \Big\{\frac{T_i+1}{2\text{Pr}(T_i=1)}-\frac{T_i-1}{2\text{Pr}(T_i=-1)}\Big\}  \Big\{Y_i- \hat{a}_0(x) -\mathcal{F}(X_i)T_i\Big\}^2.
\label{conti_aug_loss}
\end{equation}
$\mathcal{F}(X)$ is the targeted HTE, which can be estimated by any nonparametric functions, or an ensemble learner. In this article, we adopt \texttt{XGBoost} for this objective, shown later in Section \ref{tsgbt-metho}.

\subsubsection{Risk-ratio estimand for a binary response}
\label{binaryresponse}
To obtain the risk-ratio estimand of HTE as expressed in (\ref{estimand2}), we follow a similar inverse probability weighted construct and generalize a variant of the logistic loss function, which was originally proposed in Supplementary Materials of \citealp{tian2014simple}, namely
\begin{equation}
 \underset{\mathcal{F}}{\text{argmin}} \mbox{\hspace{10pt}}\mathbb{E} \Bigg[ \Big\{\frac{T+1}{2\text{Pr}(T=1)}-\frac{T-1}{2\text{Pr}(T=-1)}\Big\} \Big\{(1-Y)\mathcal{F}(X)T+ Y\exp\left(-\mathcal{F}(X)T\right)\Big\} \Bigg], \label{explossfun2}
\end{equation}
where the expectation is taken over the joint distribution of $Y, T$ and $X$. To solve for the minimizer of the loss function (\ref{explossfun2}), the estimating equation is
\begin{equation}
    \mathbb{E}\Big\{S(Y, \mathcal{F}(X)T) \Big\}=0,
    \label{weighted_bin_score}
\end{equation}
where
$S(Y, \mathcal{F}(X)T) = \bigg\{\frac{T+1}{2\text{Pr}(T=1)}+\frac{T-1}{2\text{Pr}(T=-1)}\bigg\} \bigg\{1-Y- Y\exp\left(-\mathcal{F}(X)T\right)\bigg\}$
is the estimating function from taking first-order derivative of the loss function (\ref{explossfun2}) with respect to $\mathcal{F}(X)$. We integrate the estimating equation (\ref{weighted_bin_score}) over the randomization distribution of $T$ conditional on $X$ and obtain
 $-\Big[1+\exp\big\{-\mathcal{F}(X)\big\}\Big]\mathbb{E}\left(Y|X,T=1\right) + \Big[1+\exp\big\{\mathcal{F}(X)\big\}\Big]\mathbb{E}\left(Y|X,T=-1\right) = 0$.
The minimizer is the log risk ratio estimand,
\begin{equation*}
   \mathcal{F}^{*}(x)=\log \left\{ \frac{\mathbb{E} (Y|X=x,T=1)} {\mathbb{E} (Y|X=x,T=-1)} \right\}.
\end{equation*}

To improve efficiency for estimating the risk-ratio estimand, we construct an augmented estimating equation similar to what we described in Section \ref{continuousresponse},
\begin{equation}
    \mathbb{E}\Big\{S^{\text{aug}}(Y, \mathcal{F}(X)T)\Big\}=0,
    \label{aug_weighted_bin_est_eq}
\end{equation}
where
$S^{\text{aug}}(Y, \mathcal{F}(X)T) = S(Y, \mathcal{F}(X)T) - a(X)\bigg\{\frac{T+1}{2\text{Pr}(T=1)}+\frac{T-1}{2\text{Pr}(T=-1)}\bigg\}$. The solutions of Equations (\ref{weighted_bin_score}) and (\ref{aug_weighted_bin_est_eq}) are the same because the expectation of the augmentation part is zero. Under any arbitray interaction model, any randomization ratio, and nonparametric learning of HTE, we derive one optimal augmentation term for the risk-ratio estimand, $a_0(X)= 1-\mathbb{E}(Y|X, T=1)-\mathbb{E}(Y|X, T=-1)$ in Appendix. A special case is $\text{Pr}(T=1)=\frac{1}{2}$, so that $a_0(X)= 1-2\mathbb{E}(Y|X)$. 
The optimal augmentation term $\mathbb{E}(Y|X, T=1)+\mathbb{E}(Y|X, T=-1)$ can also be estimated as one whole term through minimizing the below weighted logistic loss function,
\begin{equation}
     \underset{\mathcal{A}}{\text{argmin}}\  \mathbb{E} \Bigg[ \Big(\frac{T_i+1}{2\text{Pr}(T_i=1)}-\frac{T_i-1}{2\text{Pr}(T_i=-1)}\Big) \Big(Y_i\mathcal{A}(X_i)-\log\big(1+\exp(\mathcal{A}(X_i))\big)\Big)\Bigg].
     \label{weighted_aug_log_loss}
\end{equation}


The optimal augmentation term can be readily estimated first by minimizing the weighted logistic loss function in (\ref{weighted_aug_log_loss}), via any parametric or nonparametric estimation method. In the second stage, $\hat{a}_0(X_i)$ is plugged in the following function
\begin{equation}
    \underset{\mathcal{F}}{\text{argmin}}\  \frac{1}{n}\sum_{i=1}^n \Bigg[ \Big\{\frac{T_i+1}{2\text{Pr}(T_i=1)}-\frac{T_i-1}{2\text{Pr}(T_i=-1)}\Big\} L_i^{\text{aug}}(Y_i, \mathcal{F}(X_i)T_i)\Bigg],\label{binary_aug_loss}
\end{equation}
where $L_i^{\text{aug}}(Y_i, \mathcal{F}(X_i)T_i) = \big(1-Y_i-\hat{a}_0(X_i)\big)\mathcal{F}(X_i)T_i+Y_i\exp\Big(-\mathcal{F}(X_i)T_i\Big)$,
and $\mathcal{F}$ can be estimated by any nonparametric functions, or an ensemble learner. 


Sometimes biomarker studies in a clinical trial employs a case-control sampling scheme or more generally, retrospective sampling with additional factors. It is straightforward to incorporate the sampling weight in the estimating function by the inverse probability weighting method, as we conducted data analysis for the PCPT trial.

\subsubsection{Permutation test}
\label{permutation-metho}
One primary objective of the inference for heterogeneous treatment effects is to test whether there is any heterogeneous treatment effect at all. Permutation test has been explored for testing interactions \citep{foster2016permutation}, but one contribution of our work is formulating a permutation test for nonparametric estimation of both mean-difference and risk-ratio HTE conditional on main effects, owing to the orthogonality of the two estimation processes.  Under null hypothesis that there is no heterogeneous treatment effect, $\tau(X)$ is a constant, $\text{H}_0:\tau(X)=c$ ; under alternative hypothesis $\tau(X)$ is a function of $X$, $\text{H}_a:\tau(X)=f(X)$. The observed heterogeneous treatment effect $\tau^{\text{obs}}(x)$ is obtained from a nonparametric algorithm, for example Algorithm \ref{tsgbt_algorithm} in Section \ref{tsgbt-metho}. We calculate its sample variance across participants as $s^2\{\tau^{\text{obs}}(x)\}=\frac{1}{n-1}\sum_{i=1}^{n}\bigg(\tau^{\text{obs}}(x_i)-\tau^{\text{obs}}\bigg)^2$ with $\tau^{\text{obs}}=\frac{1}{n}\sum_{i=1}^n\tau^{\text{obs}}(x_i)$, depicting the heterogeneity of treatment effects over the distribution of $X$ \citep{levy2021fundamental}. A robust version of the sample variance can be used if there are outliers in individualized treatment effects, such as median absolute deviation. This serves as the test statistic of the following conditional permutation test: $\text{Var}_{X}\{\tau(x)\}$ is zero under $\text{H}_0$; greater than zero under $\text{H}_a$, where $\text{Var}_{X}\{\tau(x)\}$ is the variance of HTE over the distribution of X reflecting the target population. Note that other functional of $\tau(X)$ could also be tested, for example, the proportion of participants with relative risk $<$ 0.5 is greater than 0.

In the proposed permutation test, we fix the first stage estimation at $\hat{a}_0(x)$ estimated from the observed data, and we permute the rows in $x$ to obtain the permuted $x^{per}$ and estimate $\mathcal{F}(x^{per})$ using the same nonparametric algorithm used for the second stage. The expressions for the two specific loss functions with permuted data, (\ref{conti_aug_loss}) and (\ref{binary_aug_loss}), are
$\underset{\mathcal{F}}{\text{argmin}}  \mbox{\hspace{10pt}} \frac{1}{n}\sum_{i=1}^n \Big\{y_i- \hat{a}_0(x_i)-\mathcal{F}(x^{per}_i)T_i\Big\}^2,$ and
$ \underset{\mathcal{F}}{\text{argmin}}\  \frac{1}{n}\sum_{i=1}^n \bigg[\big\{1-y_i-\hat{a}_0(x_i)\big\}\mathcal{F}(x^{per}_i)t_i+y_i\exp\Big\{-\mathcal{F}(x^{per}_i)t_i\Big\} \bigg],$ with the weighting factor $\Big\{\frac{T_i+1}{2\text{Pr}(T_i=1)}-\frac{T_i-1}{2\text{Pr}(T_i=-1)}\Big\}$ ignored for simplicity. Note that $Y$ and $T$ are fixed in the permuted data, and so is the ITT effect. The next step is to obtain predicted heterogeneous treatment effects $\hat{\tau}(x^{per})$ through transforming $\mathcal{F}(x^{per})$ and compute $s^2\{\hat{\tau}(x^{per})\}$, and then repeat the permutation step sufficient number of times to obtain a permutation distribution of $s^2\{\hat{\tau}(x^{per})\}$. We define p-value as the proportion of the number of times when the estimated $s^2\{\hat{\tau}(x^{per})\}$ from the permutation distribution is equal to or greater than the observed $s^2\{\hat{\tau}^{\text{obs}}(x)\}$. The main-effect component estimated from the first stage is orthogonal to the heterogeneous treatment effect component in the second stage, therefore hypothesis testing of the latter given the estimated main effect component is valid. We evaluate our proposed permutation test for two-stage gradient boosting trees in Supplement S6 under various simulation settings.

\subsubsection{Extreme gradient boosting trees}
\label{tsgbt-metho}
In principle, any nonparametric learning algorithms can be used by our augmented learning procedure. Extreme gradient boosting trees (\texttt{XGBoost}) is particularly suitable because it allows user-defined loss functions \citep{chen2016xgboost}. Let $\mathbb{F}$ be a set of real-valued weak learner functions (i.e. basis functions), and denote similarly as in \cite{zhang2005boosting}, $\text{span}(\mathbb{F}) = \Bigg\{\sum_{m=0}^MF_m: F_m\in \mathbb{F},M\in Z^{+}\Bigg\}$,
which forms the linear function space. $F_m(X)\in\mathbb{F}$ is called the weak learner, e.g.,  classification and regression trees \citep{breiman2017classification}. Suppose $F_m(X)$ is a $J$-terminal regression tree, $F_m(x)=\omega_{q(x)}$ where $q(x):\mathbb{R}^p \to \{1,\dots, J\}$ depicts the tree structure and maps each sample to the leaf index with $J$ being the number of leaves, and $\omega_{q}:\{1,\dots, J\}\to \mathbb{R}$ denotes the weight in each leaf, such as the sample average in the leaf under mean squared error loss, with concrete forms for \texttt{XGBoost} in Supplement S3. The final output for each subject sums over the corresponding weights from each additive regression tree. Under some regularity conditions, the solution of the empirical loss minimization through classification and regression trees converges to  the minimizer of the expectation of specified loss function among $\text{span}(\mathbb{F})$ \citep{jiang2004process,zhang2005boosting,bartlett2007adaboost}.

We briefly introduce \texttt{XGBoost}\citep{chen2016xgboost}, a recent variant of GBT with regularization and faster implementation, to conduct the optimization for each additive component $F_m$. The loss function $L$ is approximated by second-order Taylor expansion with respect to $\mathcal{F}(X)$ with a regularization term. \texttt{XGBoost} proceeds in a stagewise manner. When $m=0$, let $F_0(X)$ denote the initial guess. When $m=k>0$, for the $k$-th individual additive component $F_k(X)$, we estimate it by minimizing the below loss function with all previous learned $(k-1)$ components fixed,
\begin{align}
    \mathcal{L}^{(k)}&\simeq\sum_{i=1}^n\bigg\{L\Big(Y_i,\mathcal{F}^{(k-1)}(X_i)T_i\Big)+G_iF_k(X_i)+\frac{1}{2}H_iF_k^2(X_i)\bigg\} + \gamma J_k
    \label{kthtreeloss}\\
    &= \sum_{j=1}^{J_k}\bigg\{\Big(\sum_{i\in I_j}G_i\Big)\omega_j+\frac{1}{2}\Big(\sum_{i\in I_j}H_i\Big)\omega_j^2\bigg\} + \gamma J_k, \label{kthtreeloss2}
\end{align}
where $G_i=\partial L/\partial F_k|_{F_k=0}$ and $H_i=\partial^2 L/(\partial F_k)^2|_{F_k=0}$, which are the respective gradient and hessian functions of the loss function for the $i$-th sample, $\mathcal{F}^{(k-1)}(X_i)=\sum_{m=0}^{k-1}\widehat{F}_m(X_i),$ $J_k$ denotes the number of leaves for the $k$-th regression tree and $\gamma$ is the regularization parameter, penalizing the complexity of tree structure (i.e. the minimum loss reduction required to make further partition). For the second equation, we write out the approximated loss function (\ref{kthtreeloss}) by removing the constant $L\Big(Y_i,\mathcal{F}^{(k-1)}(X_i)T_i\Big)$ in terms of the fixed tree structure $q_k(X)$ where $I_j=\{i|q_k(x_i)=j\}$ denotes the set for $j$-th leaf in the $k$-th tree.

\texttt{XGBoost} allows a customized loss function where we only need to provide the gradient and hessian functions of the specified loss function. Thus, it provides one off-the-shelf learning technique to implement our augmented learning procedure for estimating $\mathcal{F}(X)$. We implement \texttt{XGBoost} for estimating both the augmentation term $a_0(X)$ and the function of HTE $\mathcal{F}(X)$ in a two-stage fashion, shown by Algorithm \ref{tsgbt_algorithm} in Appendix, called Two-Stage Gradient Boosting Trees (TSGBT). We note that the first stage augmentation term does not have to be estimated by GBT -- any parametric and nonparametric algorithm should work. In the algorithm, denote by $M_a$ and $M$ the total number of additive regression trees (i.e., number of iterations) for the respective two stages. $G, G^a$ and $H, H^a$ are the first- and second-order gradient statistics of any convex loss functions. Let $J_{ak}$ and $J_k$ denote the number of leaves for $k$-th regression tree.

First, we apply \texttt{XGBoost} algorithm for estimating $a_0(X)$ by specifying its corresponding gradient $G_i^{a}$ and hessian functions $H_i^{a}$ for a continuous response with mean squared error loss or a binary response with logistic likelihood loss.
Second, we connect the $k$-th loss function $\mathcal{L}^{(k)}$ to our proposed augmented loss functions through its first- and second-order gradient statistics, the loss function (\ref{conti_aug_loss}) in Section \ref{continuousresponse} and (\ref{binary_aug_loss}) in Section \ref{binaryresponse} with their respective augmentation term $a_0(X_i)$ estimated in the first stage. For estimating the $k$-th regression tree,
$G_i=-2T_i\bigg[Y_i-\widehat{\mathbb{E}}(Y_i|X_i)-\mathcal{F}^{(k-1)}(X_i)T_i\bigg]$ for loss  function (\ref{conti_aug_loss}), or
$-T_i\bigg[Y_i-1-2\widehat{\mathbb{E}}(Y_i|X_i)+1+Y_i\exp\Big(-\mathcal{F}^{(k-1)}(X_i)T_i\Big)\bigg]$ for loss function (\ref{binary_aug_loss}); $H_i=2$ for loss function (\ref{conti_aug_loss}), or
$Y_i\exp\Big(-\mathcal{F}^{(k-1)}(X_i)T_i\Big)$ for loss function (\ref{binary_aug_loss}). When the randomization ratio is unequal, all the $G_i$ and $H_i$ should multiply one weight factor $\Big\{\frac{T_i+1}{2\text{Pr}(T_i=1)}-\frac{T_i-1}{2\text{Pr}(T_i=-1)}\Big\}$.
After $M$ number of iterations, $\mathcal{F}^{(M)}(X)$ is the estimator for $\mathcal{F}(X)$.

Boosting algorithm can be overfitting the data, and proper regularization is needed to protect the consistency property\citep{buhlmann2003boosting,jiang2004process,zhang2005boosting,buehlmann2006boosting}. We shrink the learned additive component $A(X)$ or $F(X)$ at each iteration by a learning rate, denoted as $\eta$. The number of iterations and the learning rate have a trade-off relationship. Boosting trees with a smaller learning rate learn slower, requiring a larger number of iterations. Yet a small learning rate commonly guarantees better convergence property \citep{zhang2005boosting}. The hyper-parameters are crucial for the performance of prediction in gradient boosting trees. Other than the regularization parameter $\gamma$ and the learning rate $\eta$, \texttt{XGBoost} has several other tuning parameters: the number of iterations $M$, maximum depth of a tree (\emph{max-depth}), row sampling rate (\emph{subsample}), column sampling rate (\emph{colsample}), minimum sum of instance weight(i.e. the second-order derivative of the specified loss function for each sample) needed in a child leaf (\emph{min-child-weight}). We describe the tuning procedure for these parameters in Supplement S2.


\section{Numerical Studies}
\label{sec:simulations}
In this section, we evaluate the proposed two-stage approach for estimating heterogeneous treatment effects and the proposed permutation test for no interaction effect under different scenarios of simulations. For brevity, we present the selected representative results and leave additional results to Web Appendix B, including a simple proof-of-principle simulation experiment to show the consistency and robustness of the two-stage procedure. 

We compare the proposed two-stage gradient boosting trees (TSGBT) approach with full regression (FR), the modified covariates approach (MC), the  modified covariates with efficiency augmentation (MC-aug), separate GBT and weighted GBT. Full regression is a multivariate linear model with the outcome $Y$, the covariates $X$, the treatment indicator $T$, and the interaction from the covariates and the treatment indicator $XT$. The modified covariates method is a linear model with the outcome $Y$ and the modified covariates $XT$ \citep{tian2014simple}. The modified covariates with efficiency augmentation is a linear model with the outcome $Y$, the modified covariates $XT$, and one augmentation term. For these parametric approaches, FR, MC, MC-aug, we conduct the variable selection and reguarization through LASSO, using the {\tt cv.glmnet} function in {\tt R} package {\tt glmnet}. The separate GBT (SGBT) approach is fitting GBT to two treatment groups separately. The weighted GBT is the second stage of our proposed two-stage GBT without the first stage. We conduct gradient boosting trees by using functions in packages {\tt XGBoost} and {\tt caret} with our proposed modified loss functions. We assess the performance of these approaches through spearman correlation (sCORR) and mean squared error (MSE).

\subsection{Continuous outcome}
\label{continuousoutcome-est}
Assume $Y$ is continuous, we generate N = 300 independent Gaussian samples from the regression model,
$Y=(\alpha_0+\sum_{j=1}^p\alpha_jX_j)^2+\mathcal{F}(X;\boldsymbol{\beta})\cdot T/2+\sigma_0\epsilon,$
where $\mathcal{F}(X;\boldsymbol{\beta})$ can be any functions, the covariates $(X_1,\dots,X_p)\sim \mathcal{N}(\boldsymbol{0}, \Sigma)$ with $\Sigma$ being a first-order auto-regressive variance-covariance matrix with correlation $\rho=0.5$, $p=50 \text{ or } 1000$, $\sigma_0=2$ and $\epsilon \sim \mathcal{N}(0,1)$. The treatment indicator $T$ is randomly assigned with probability 1/2 for both $T=1$ and $T=-1$. We first consider the scenario with linear interaction effects,
$\mathcal{F}(X;\boldsymbol{\beta})=\beta_0+\sum_{j=1}^p\beta_jX_j.$ Four parameter settings were designed as follows. In {\bf Setting 1}, $(\alpha_0,\alpha_1,\alpha_2,\alpha_3,\alpha_4,\alpha_5,\dots,\alpha_p)=(0.4,0.6,$-$0.6,0.6,0.6,0,\dots,0)$, $(\beta_0,\beta_1,\beta_2,\beta_3,\beta_4, \beta_5,\dots,\beta_p)=(0.8,0.8,$-$0.8,0.8,0.8,0,\dots,0).$ The goal is to assess how the proposed tree-boosting method perform in simple linear models, where other methods are able to capture the true model. We then consider the three other settings with more complex interaction effects. Let $\alpha_0=(\sqrt{3})^{-1}, \alpha_j=(2\sqrt{3})^{-1}, j=3,4,\dots, 10,$ with others being zeros and $\mathcal{F}(X;\boldsymbol{\beta})=\beta_0+\sum_{j=1}^p\beta_j(X_j+X_j^2)+\sum_{1\leq i <j\leq p}\beta_{ij}X_iX_j.$ In {\bf Setting 2}, $(\beta_0,\beta_1,\beta_2,\beta_3,\beta_4,\beta_5,\dots, \beta_p)=(0.8,1.6,$-$1.6,1.6,$-$1.6,0,\dots,0), \beta_{12}=1.6, \beta_{15}=1.6$. In {\bf Setting 3},
    $(\beta_0,\beta_1,\beta_2,\beta_3,\beta_4,\beta_5, \dots,\beta_p)=(0.8,0.8,$-$0.8,0.8,$-$0.8,0,\dots,0), \beta_{12}=0.8, \beta_{15}=0.8$.
In {\bf Setting 4},
   $(\beta_0,\beta_1,\beta_2,\beta_3,\beta_4, \beta_5,\dots,\beta_p)=(0.8,0,0,0,0,0,\dots,0), \beta_{12}=0, \beta_{15}=0$.
Setting 2 and 3 are proposed to illustrate performance of boosting tree algorithms under strong and moderate interaction effects. Setting 4 is used to evaluate how these different methods perform when there is no interaction effects. We compared all methods under the four settings with p=50 or 1000.

\begin{figure}
\centering
\subfigure[sCORR, Setting 1]{%
\includegraphics[width=0.3\linewidth, height=5cm]{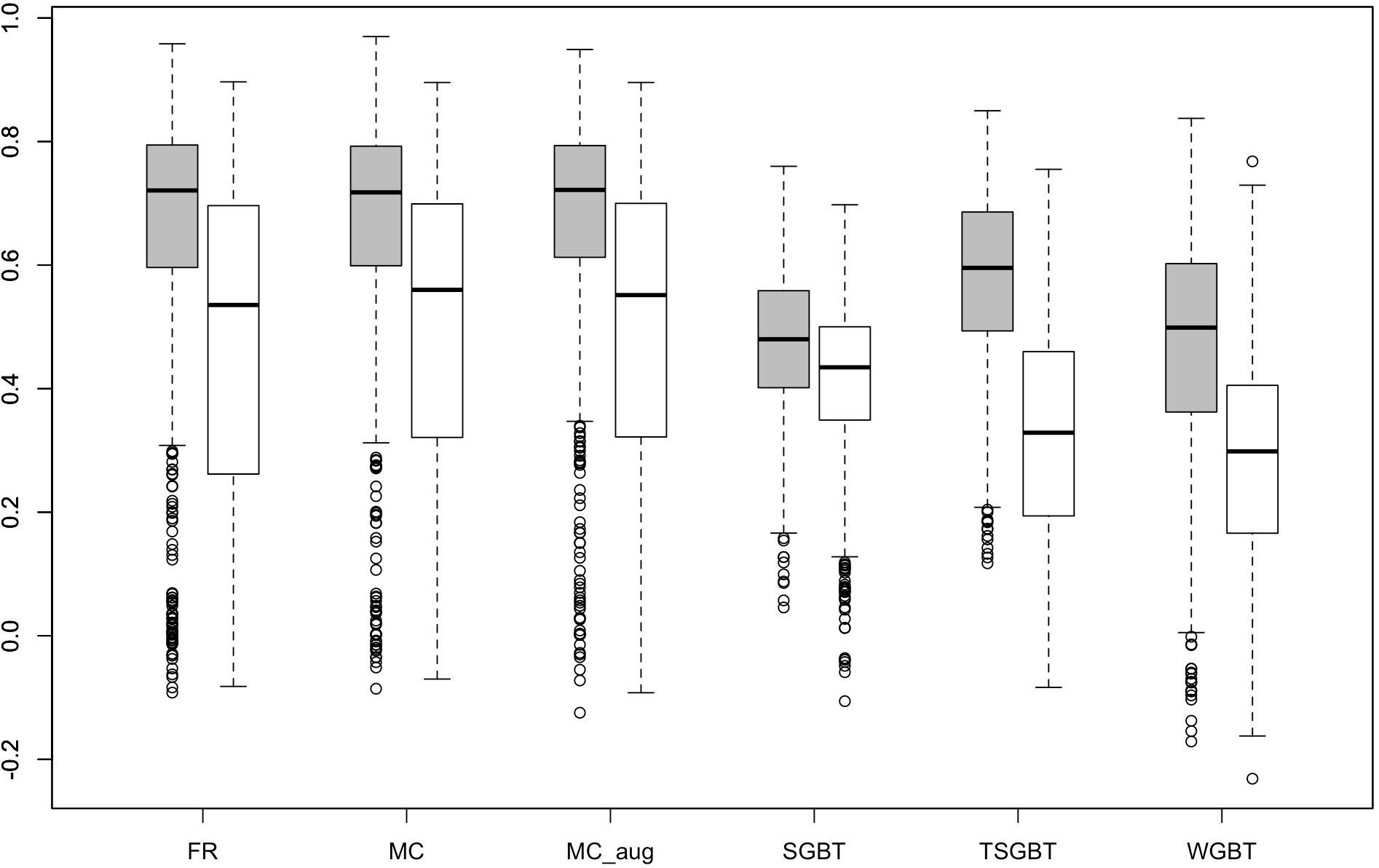}
\label{fig:conti1scorr}}
\quad
\subfigure[sCORR, Setting 2]{%
\includegraphics[width=0.3\linewidth, height=5cm]{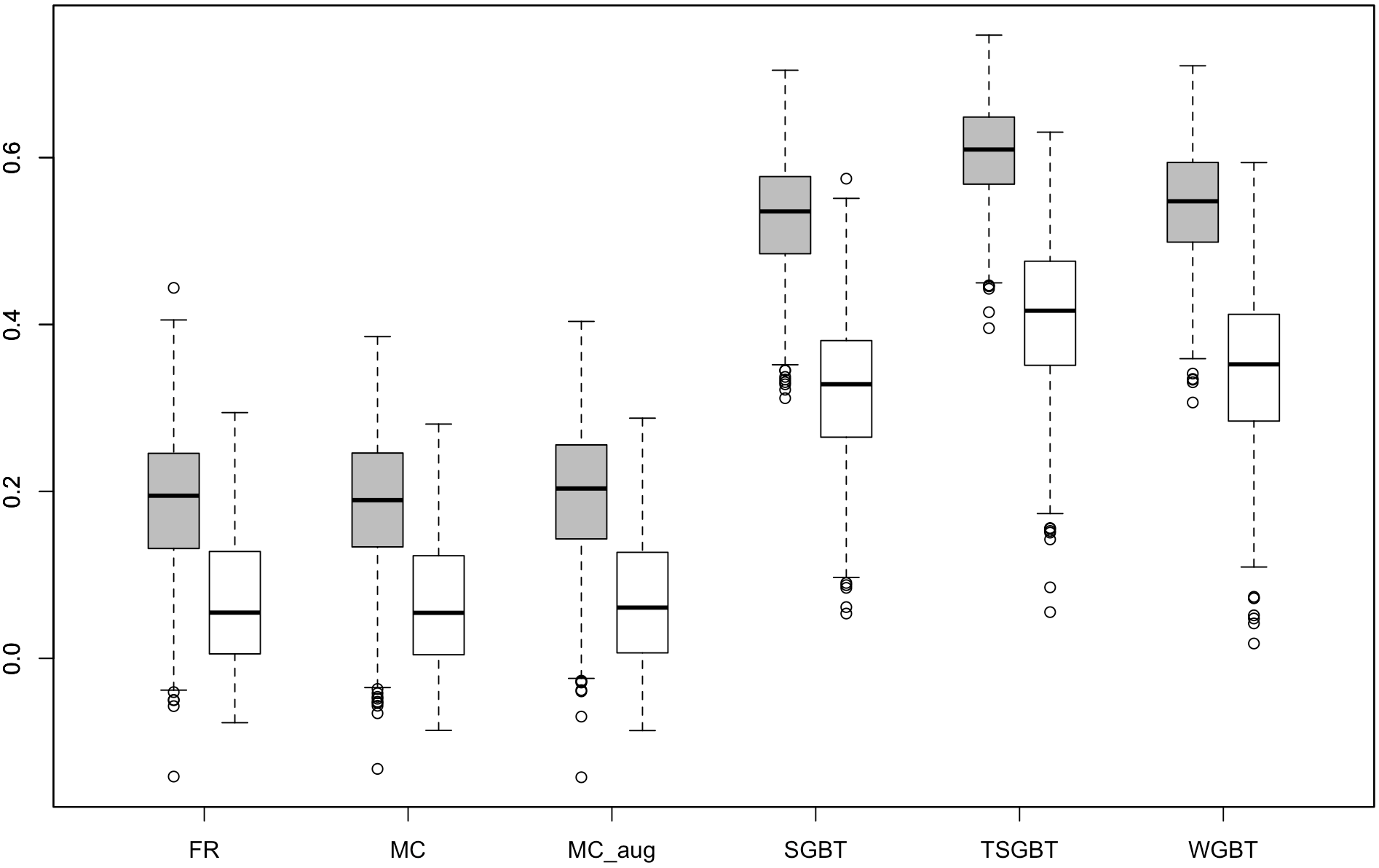}
\label{fig:conti2scorr}}
\quad
\subfigure[sCORR, Setting 3]{%
\includegraphics[width=0.3\linewidth, height=5cm]{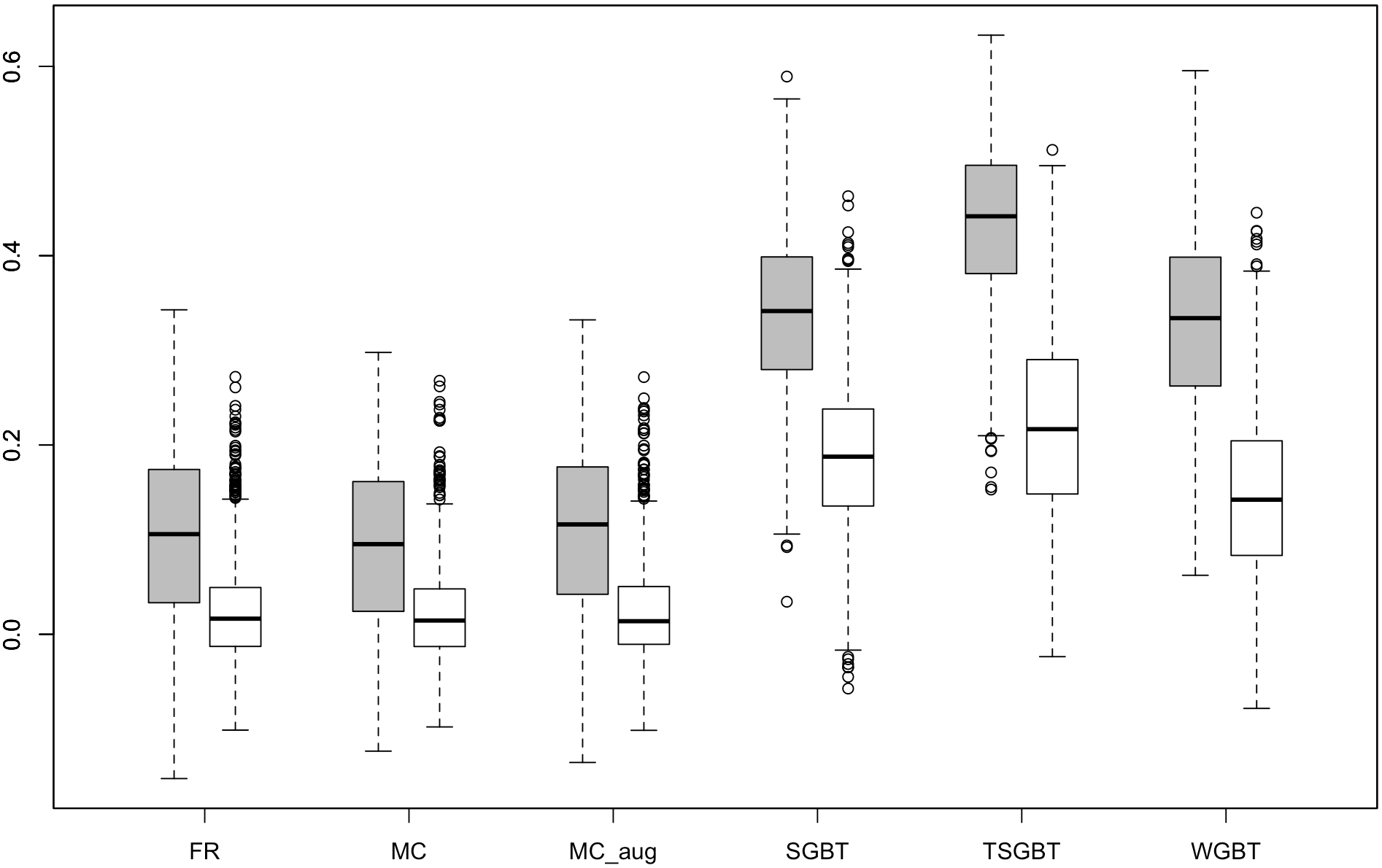}
\label{fig:conti3scorr}}
\subfigure[MSE, Setting 1]{%
\includegraphics[width=0.3\linewidth, height=5cm]{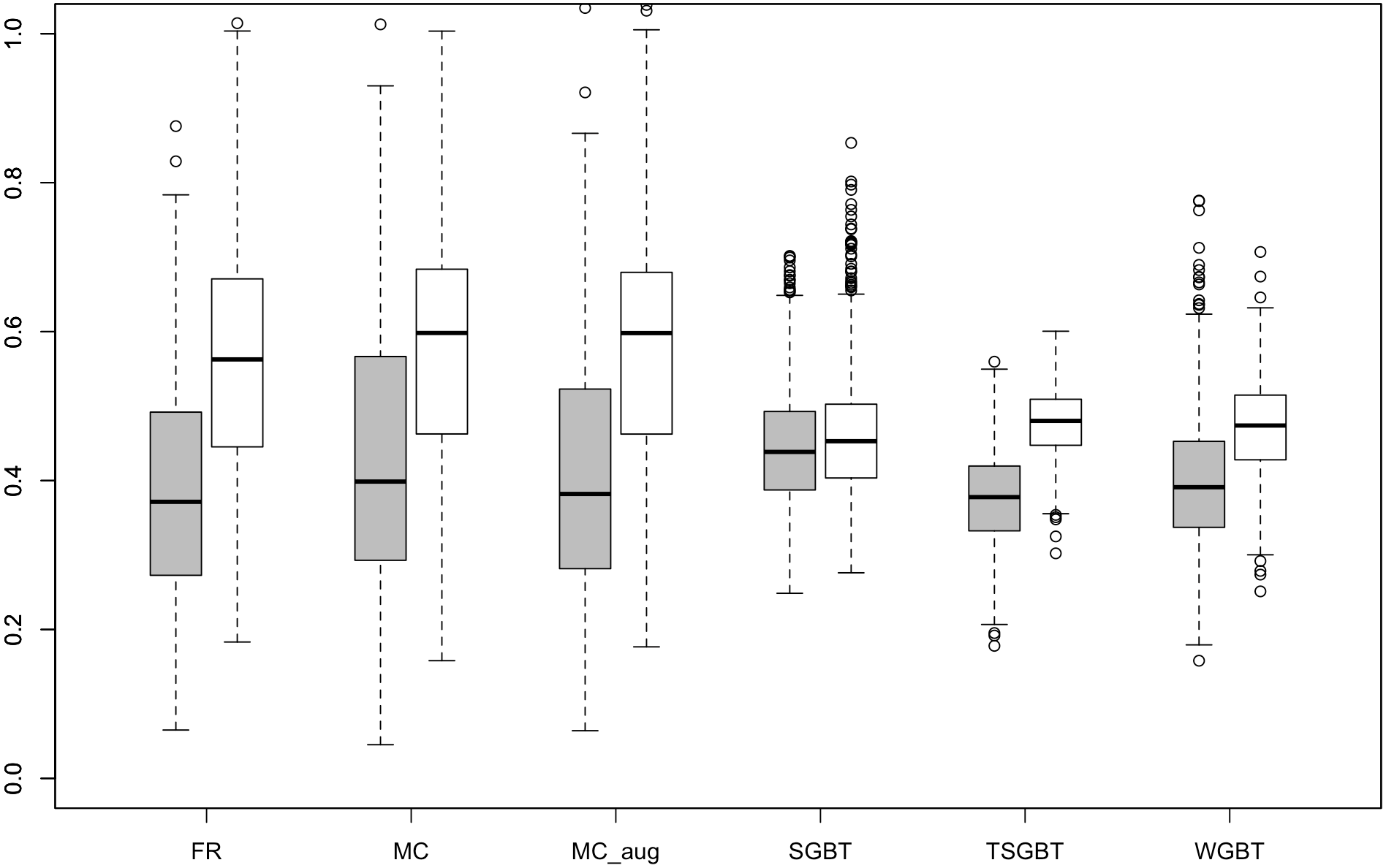}
\label{fig:conti1mse}}
\quad
\subfigure[MSE, Setting 2]{%
\includegraphics[width=0.3\linewidth, height=5cm]{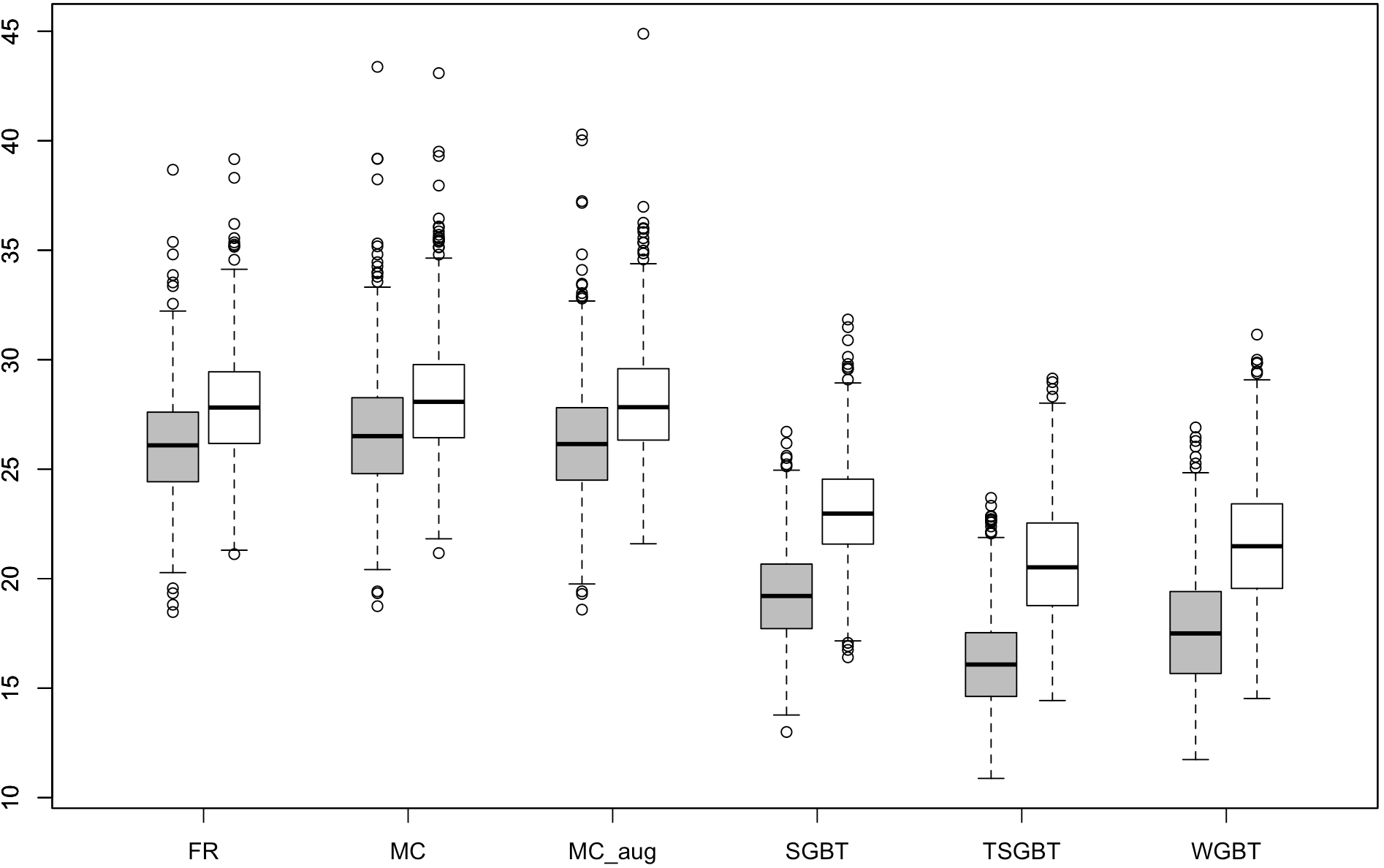}
\label{fig:conti2mse}}
\quad
\subfigure[MSE, Setting 3]{%
\includegraphics[width=0.3\linewidth, height=5cm]{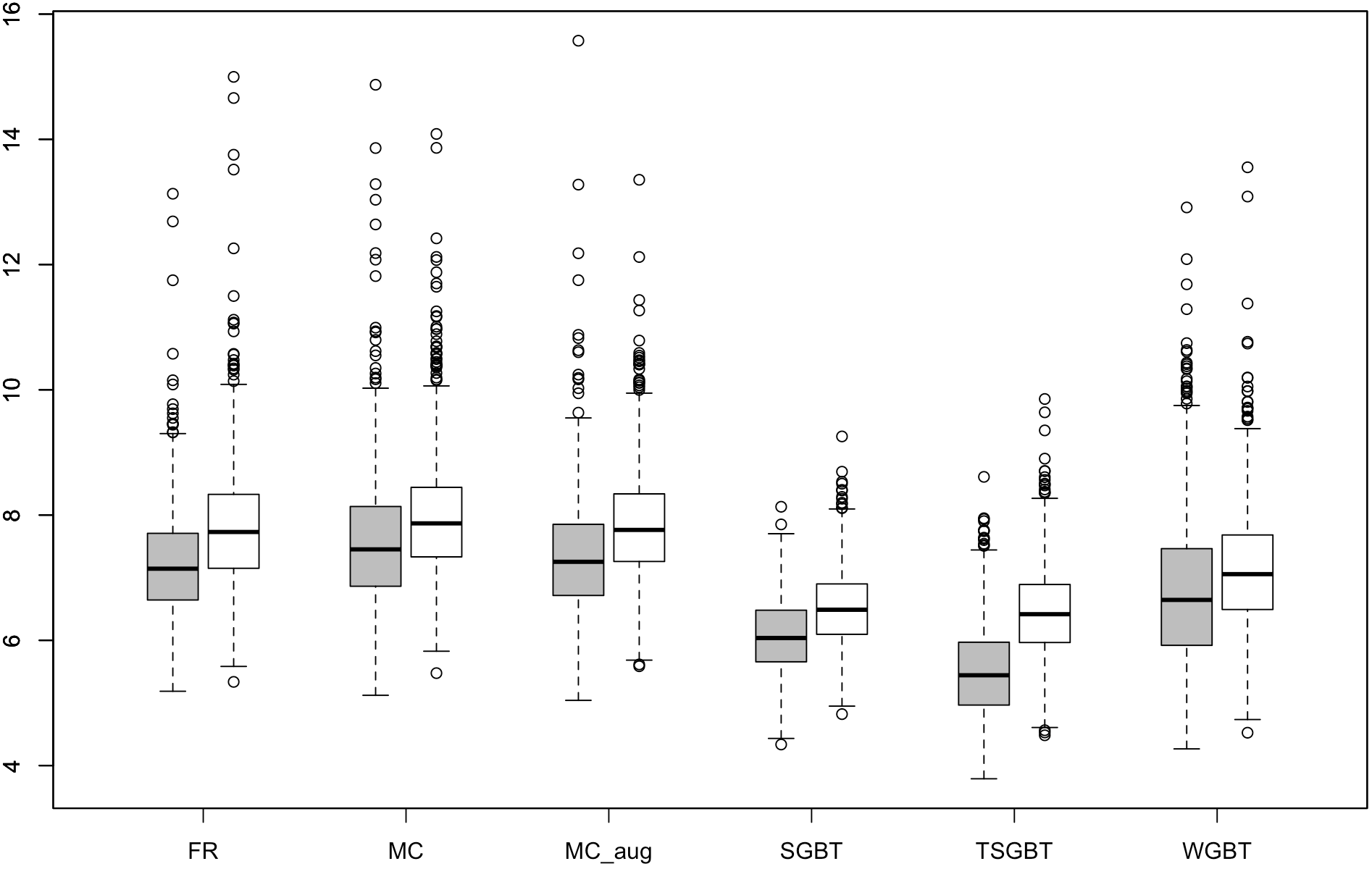}
\label{fig:conti3mse}}

\subfigure[MSE, Setting 4]{%
\includegraphics[width=0.3\linewidth, height=5cm]{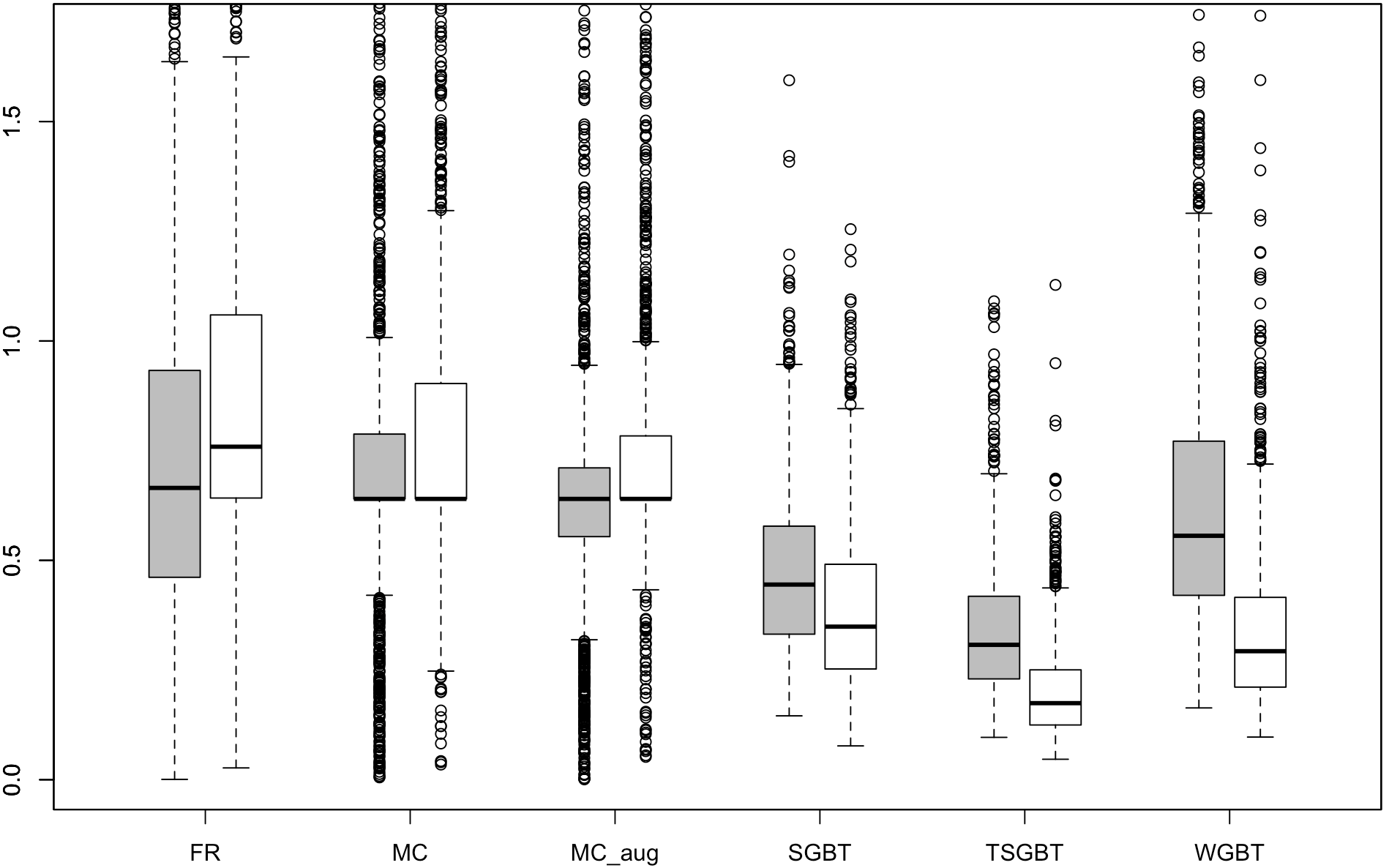}
\label{fig:conti4mse}}
\quad
\subfigure[Diagnostic Plot for TSGBT, Setting 4 with p = 50]{%
\includegraphics[width=0.3\linewidth, height=5cm]{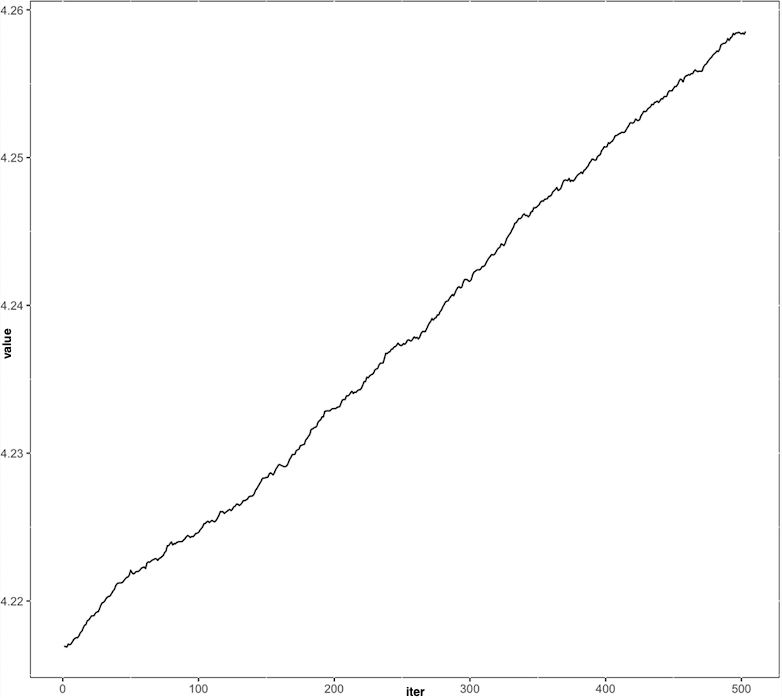}
\label{fig:conti4diagp50}}
\quad
\subfigure[Diagnostic Plot for TSGBT, Setting 4 with p = 1000]{%
\includegraphics[width=0.3\linewidth, height=5cm]{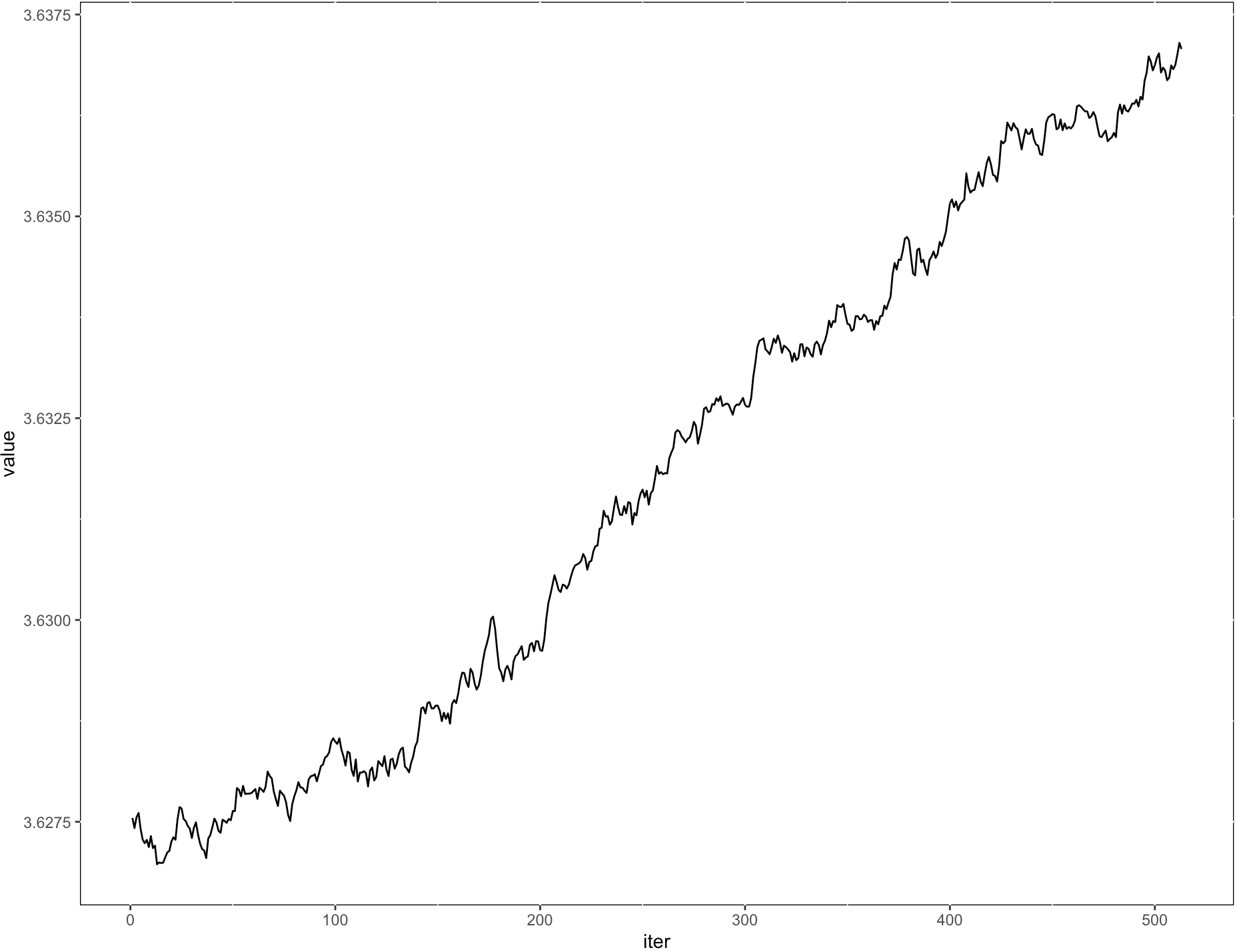}
\label{fig:conti4diagp1000}}
\caption{Boxplots of the rank correlation coefficients and mean squared errors between the estimated heterogeneous treatment effects and the true effects using six different methods (FR, full regression method; MC, modified covariates; MC-aug, modified covariates with efficiency augmentation method; SGBT, separate gradient boosting trees; WGBT, weighted gradient boosting trees; TSGBT, two-stage gradient boosting trees), when applied to simulated data with Gaussian outcomes. The grey and white boxes represent low- and high-dimensional (p=50 and 1000) settings, respectively. Results for Setting 1-3 were shown in (a)-(f) and the results for Setting 4 were shown in (g)-(i). (h) and (i) show the diagnostic plots of the 2nd stage GBT for selecting best number of iterations with the rooted mean squared error values of internal testing data being the vertical axis.}
\label{fig:continuous}
\end{figure}

The true heterogeneous treatment effect is, $\tau(X)=\text{E}[Y|X,T=1]-\text{E}[Y|X,T=-1]=\mathcal{F}(X;\boldsymbol{\beta})$.
For each simulation, we generated 1000 independent testing data to evaluate the performance of these approaches. In Figure \ref{fig:continuous}, the top three panels are boxplots of the spearman correlations for Settings 1, 2 and 3, respectively; the middle three panels are the MSE plots for Settings 1, 2 and 3, respectively; panel (h) is the MSE plot under Setting 4; panel (i) and (g) are the diagnostic plots to tune the best number of iterations under p = 50 and 1000, respectively. The grey boxes are for the low-dimensional settings, while the white boxes for high-dimensional settings.

In Figure \ref{fig:continuous}, TSGBT performed uniformly better than WGBT across all settings, confirming that the first stage augmentation improves the efficiency for estimating HTE. When the number of covariates increases from 50 to 1000, almost all sCORR decrease and MSE increase, because high dimensional covariates introduce more noise and render estimation more difficult. Figure \ref{fig:conti1scorr} shows that, when the true models are linear models, all linear methods (FR, MC, MC-aug) performed  better than boosting approaches regarding the ranking of estimated treatment effects. For MSE, all boosting methods were comparable to or even better than linear methods (Figure \ref{fig:conti1mse}). For Setting 2 and 3, Figures \ref{fig:conti2scorr}, \ref{fig:conti3scorr}, \ref{fig:conti2mse} and \ref{fig:conti3mse} show that boosting methods performed better than the linear methods because flexible tree structures can better model more complicated non-linear interaction effects; the linear models using LASSO could not successfully select the covariates. TSGBT was better than SGBT under Settings 2 and 3 with both low- and high-dimensional cases, because TSGBT makes use of all the data to estimate the main effects and the interaction effects for both groups, while SGBT splits the data for two treatment groups, leading to the loss of efficiency. For example, the median values for sCORR under Setting 2 were 0.20(0.06), 0.19(0.06), 0.20(0.06), 0.54(0.33), 0.61(0.42), 0.55(0.35)  for FR, MC, MC-aug, SGBT and TSGBT under p = 50(1000), respectively; the median values for MSE under Setting 2 were 26.09(27.82), 26.51(28.08), 26.15(27.83), 19.21(22.97), 16.08(20.52), 17.50(21.48) for FR, MC, MC-aug, SGBT and TSGBT under p = 50(1000), respectively.

Figures \ref{fig:conti4diagp50} and \ref{fig:conti4diagp1000} shows that when there is no interaction effect at all, GBT diagnostic plots can immediately show the null HTE. These diagnostic plots adds the benefit of TSGBT versus SGBT: the former is able to use the GBT machinary to evaluate the signal-to-noise ratio, while the latter cannot. We also compared the performance of all these methods under Settings 2 and 3 when the correlation is zero and when the covariance matrix follows a compound symmetric matrix with $\rho=1/3$ (results shown in Web Appendix B). The conclusions follow similarly.

\subsection{Binary outcome}
\label{binaryoutcome-est}
When $Y$ is binary, we generate N = 1500 independent binary samples from the exponential model for relative risk, $\log{\frac{P_1}{P_{-1}}}=2\mathcal{F}(X;\boldsymbol{\beta})$ with its nuisance model for the odds product, $\log{\frac{P_1P_{-1}}{(1-P_1)(1-P_{-1})}}=-C-(\alpha_0+\sum_{j=1}^p\alpha_jX_j)^2$ \citep{richardson2017modeling}, where both $P_1=\mathbb{E}[Y|X,T=1]$ and $P_{-1}=\mathbb{E}[Y|X,T=-1]$ are derived from the specified primary and nuisance models, and
$C$ is chosen to guarantee approximately the prevalence rate varying between 0.2 and 0.3, and the proportion of non-negative exponents less than 0.01. Additional ways to simulate binary outcomes were left in Web Appendix B. We consider the following three settings of simulations under complex interaction effects scenario. Specifically, Settings 1, 2 and 3 are proposed to illustrate performance of boosting tree algorithms under strong, moderate and no interaction effects, respectively. Let $(\alpha_0, \alpha_1, \alpha_2, \alpha_3, \alpha_4, \alpha_4, \dots, \alpha_p)=(0.4, 0.6, $-$0.6, 0.6, 0.6, 0, \dots, 0)$ and $\mathcal{F}(X;\boldsymbol{\beta})=\beta_0+\sum_{j=1}^p(\beta_jX_j+\gamma_jX_j^2)+\sum_{1\leq i <j\leq p}\beta_{ij}X_iX_j.$ In {\bf Setting 1}, $(\beta_0,\beta_1,\beta_2,\beta_3,\beta_4,\beta_5,\dots,\beta_p)=(0.3,0.3,0.4,0.3,0.4,0,\dots,0), (\gamma_1,\gamma_2,\gamma_3,\gamma_4,\gamma_5,\dots,\gamma_p)=(0.4,$-$0.4,0.4,0.4,0,\dots,0), \beta_{12}=0.5, \beta_{15}=0.5$, $C=2.5$;
In {\bf Setting 2},
$(\beta_0,\beta_1,\beta_2,\beta_3,\beta_4,\beta_5,\dots,\beta_p)=(0.1,0.1,0.2,0.1,0.2,0,\dots,0), (\gamma_1, \gamma_2,\gamma_3,\gamma_4,\gamma_5,\dots,\gamma_p)=(0.2,$-$0.2,0.2,0.2,0,\dots,0), \beta_{12}=0.5, \beta_{15}=0.5$, $C=2.5$; In {\bf Setting 3}, $(\beta_0,\beta_1,\dots,\beta_p)=(0.3,0,\dots,0), (\gamma_1,\dots,\gamma_p)=(0,\dots,0), \beta_{12}=0, \beta_{15}=0$, $C=2$.

\begin{figure}
\centering
\subfigure[sCORR, Setting 1]{%
\includegraphics[width=0.45\linewidth, height=5cm]{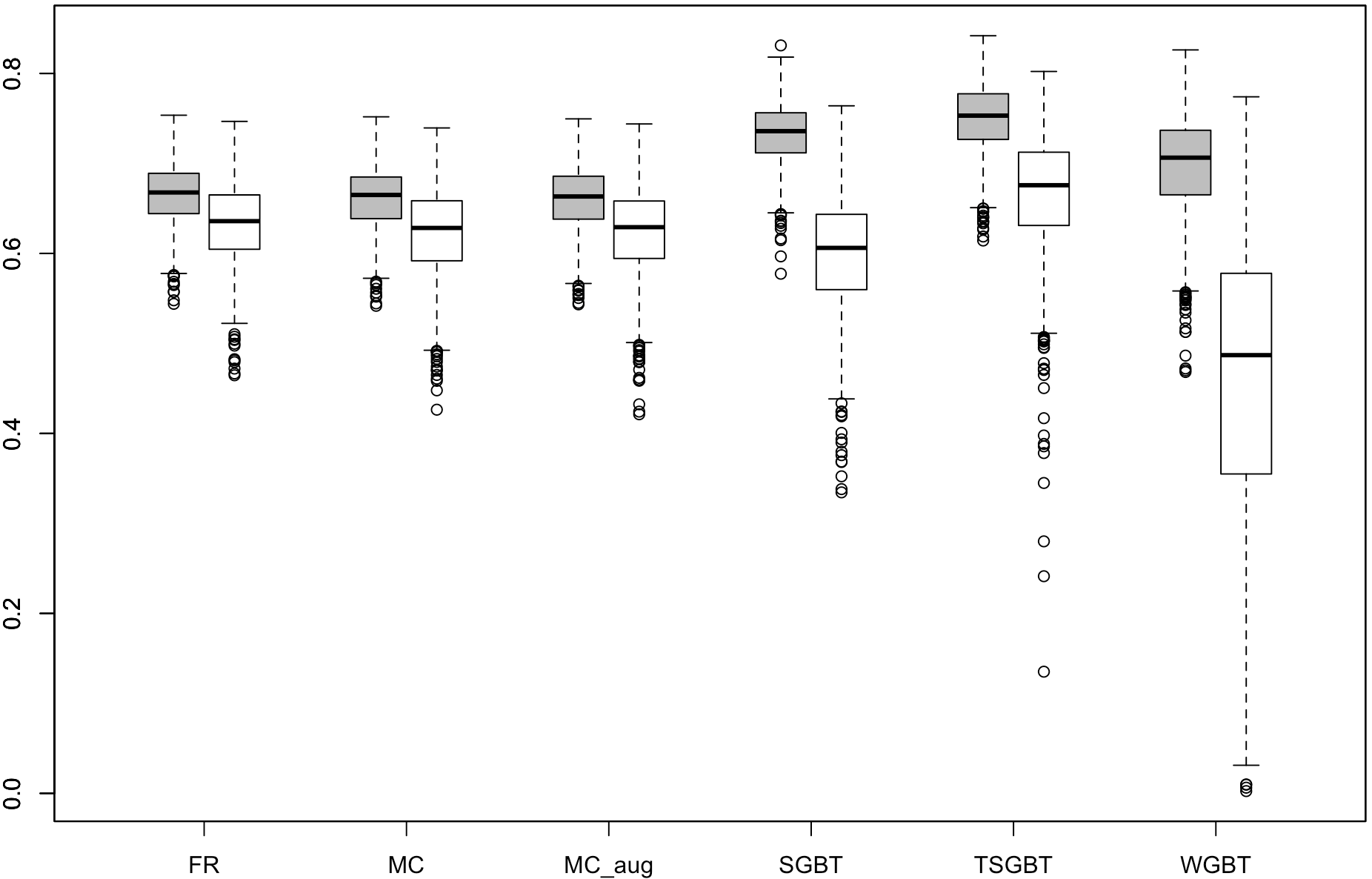}
\label{fig:binary1scorr}}
\quad
\subfigure[sCORR, Setting 2]{%
\includegraphics[width=0.45\linewidth, height=5cm]{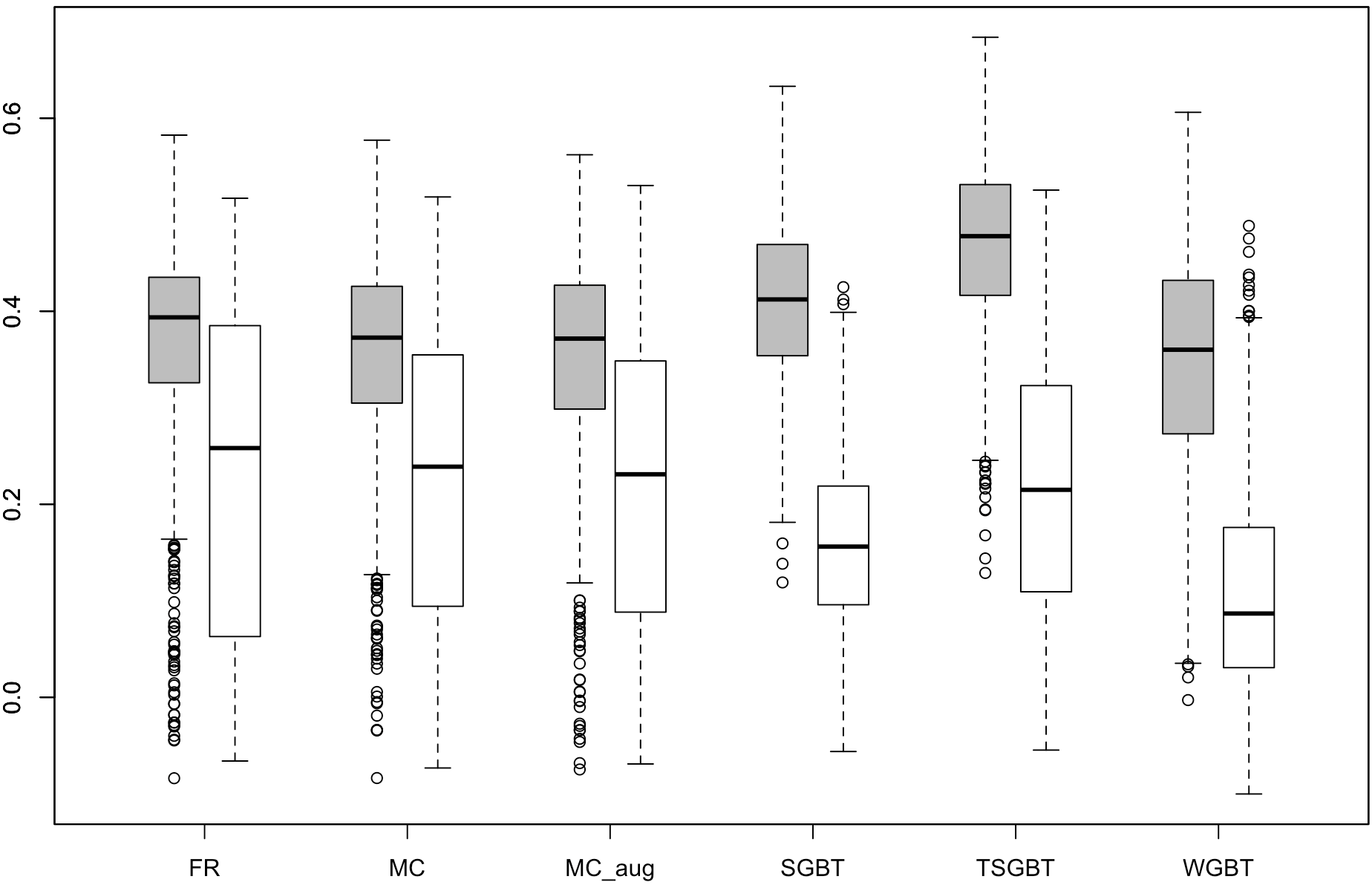}
\label{fig:binary2scorr}}
\subfigure[MSE, Setting 1]{%
\includegraphics[width=0.45\linewidth, height=5cm]{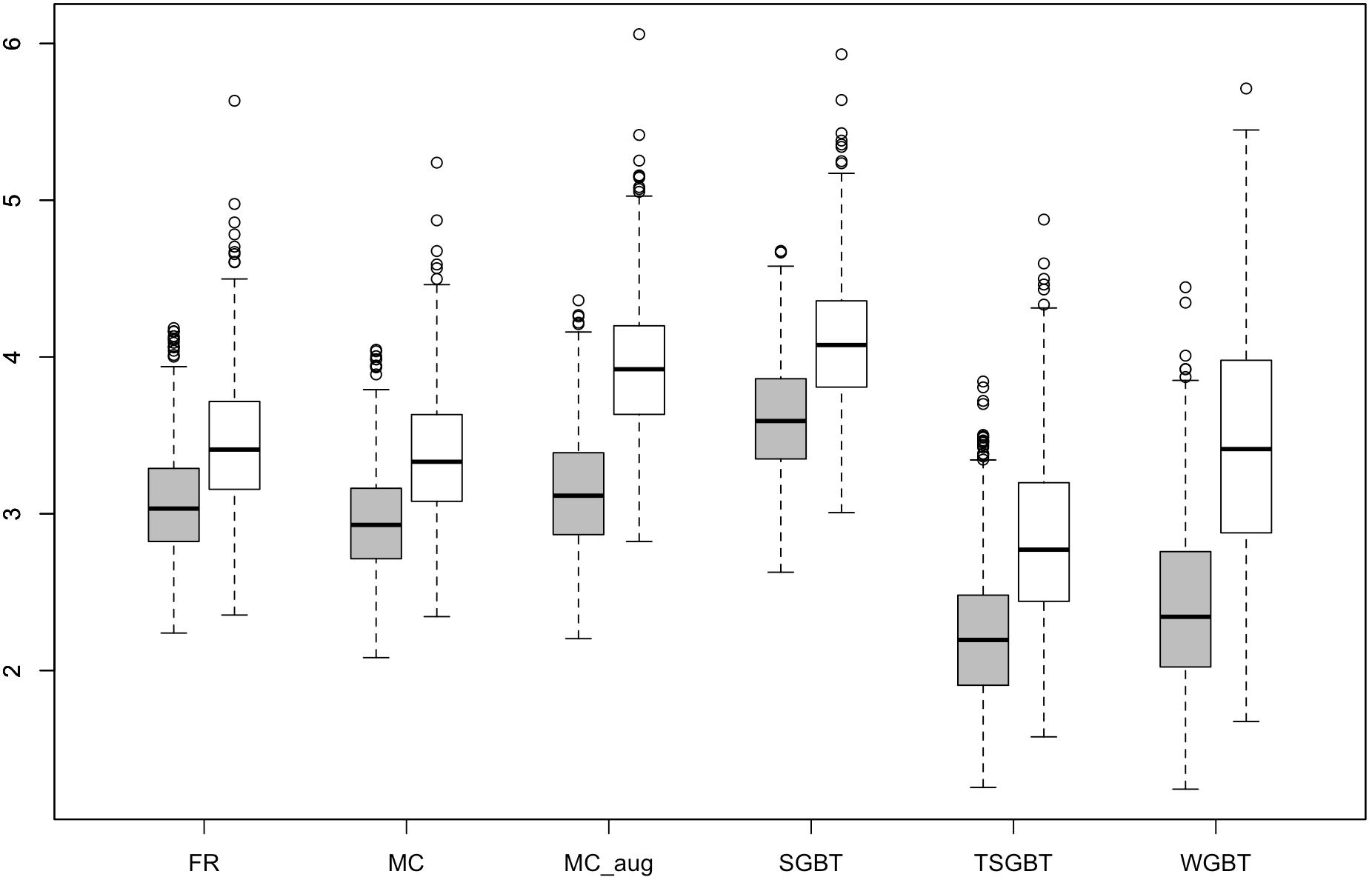}
\label{fig:binary1mse}}
\quad
\subfigure[MSE, Setting 2]{%
\includegraphics[width=0.45\linewidth, height=5cm]{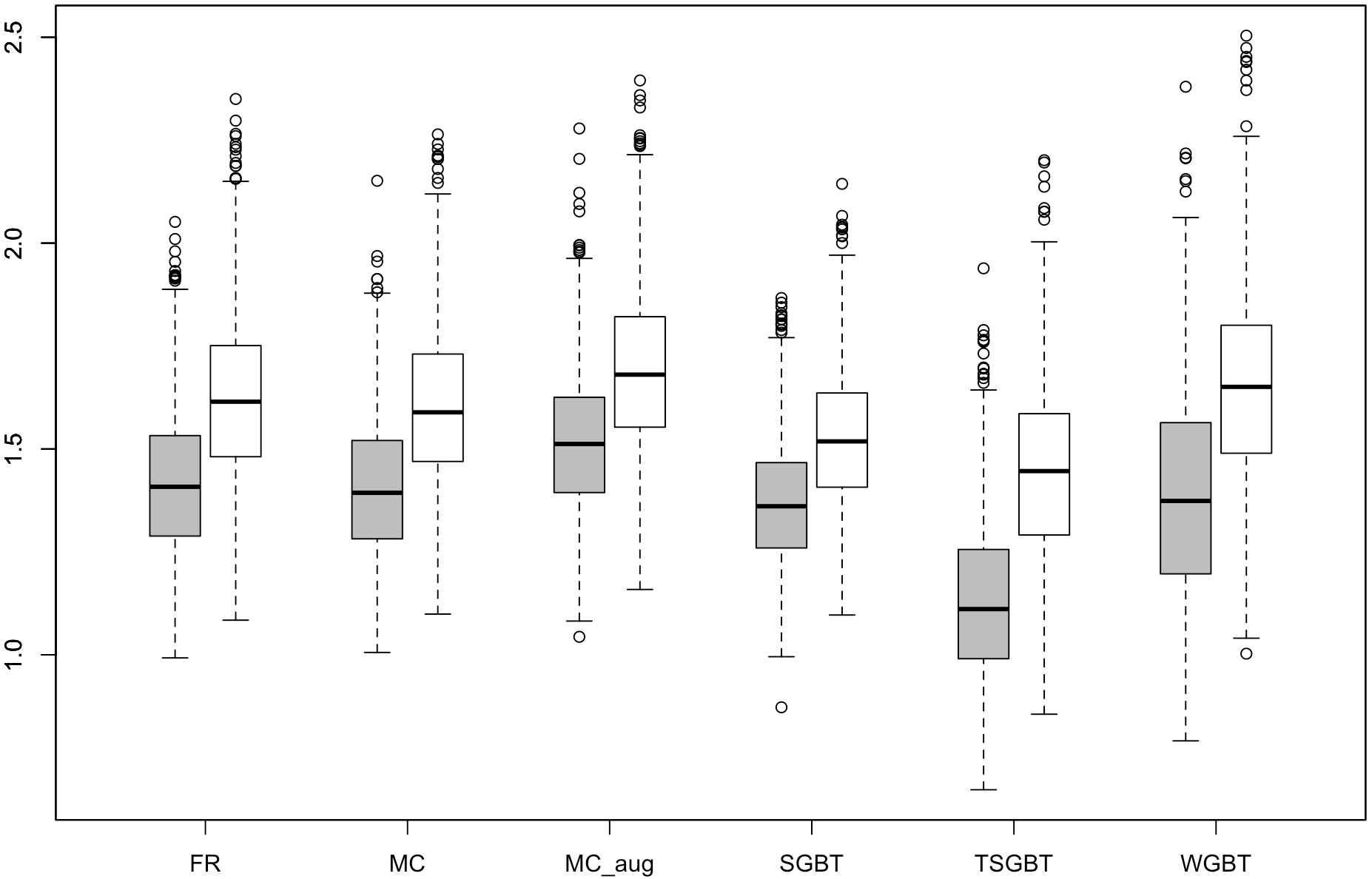}
\label{fig:binary2mse}}
\subfigure[MSE, Setting 3]{%
\includegraphics[width=0.45\linewidth, height=5cm]{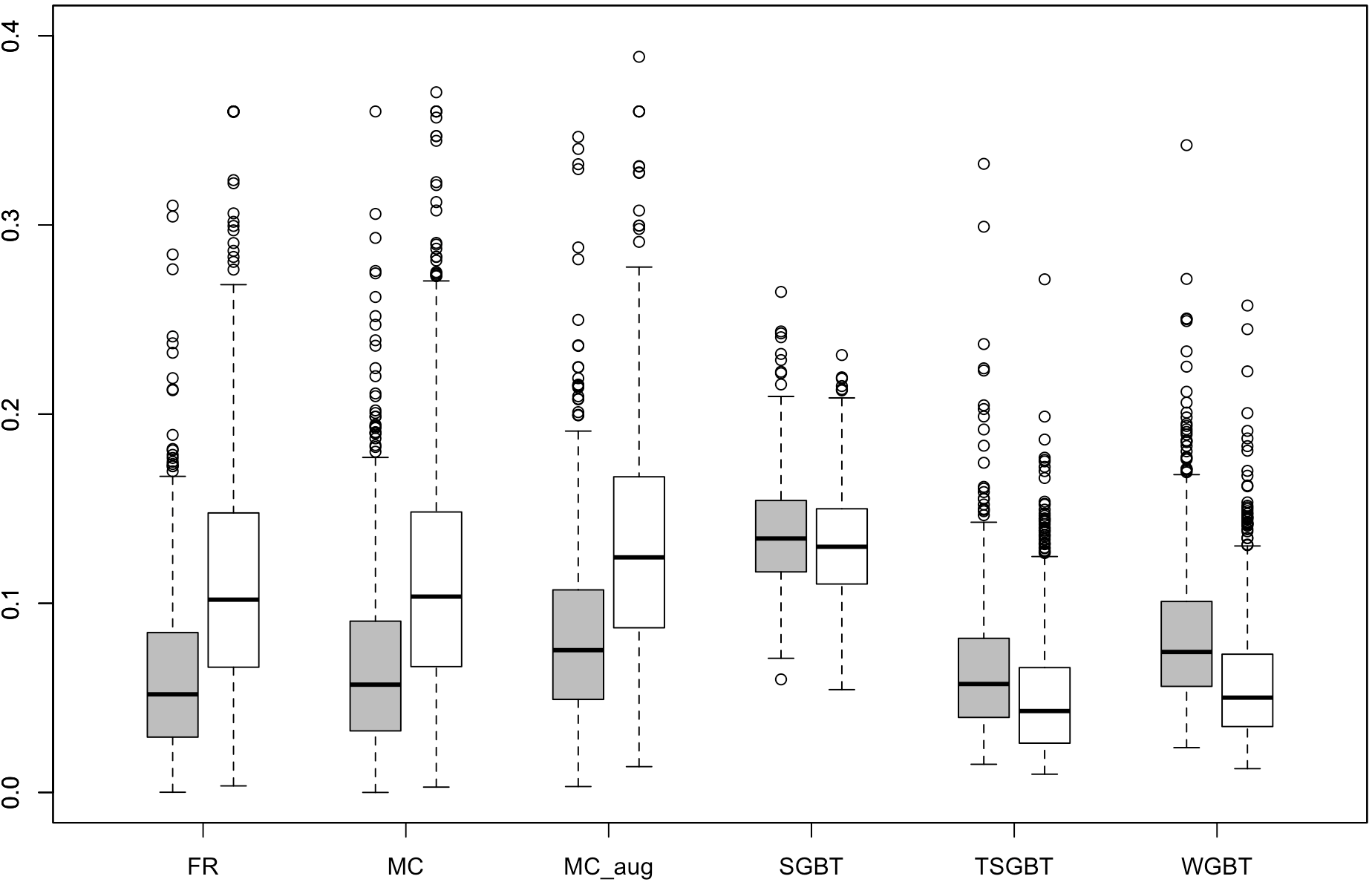}
\label{fig:binary3mse}}

\caption{Boxplots of the rank correlation coefficients and mean squared errors for Settings 1, 2, and 3 between the estimated heterogeneous treatment effects and the true effects with six different methods (FR, full regression method; MC, modified covariates; MC-aug, modified covariates with efficiency augmentation method; SGBT, separate gradient boosting trees; WGBT, weighted gradient boosting trees; TSGBT, two-stage gradient boosting trees) applied to binary outcomes. The grey and white boxes represent low- and high-dimensional (p=50 and 1000) cases, respectively.}
\label{fig:binary}
\end{figure}

We applied the six methods to the generated binary data under the three settings with p=50 or 1000. The true heterogeneous treatment effect is defined in terms of risk ratio scale, $\tau(X)=\frac{\mathbb{E}[Y|X,T=1]}{\mathbb{E}[Y|X,T=-1]}$.
We compared the estimated scores with the true values through sCORR and MSE, where MSE is calculated in the log risk ratio scale. We generated 1000 independent testing data to evaluate the performance of these approaches. The results  in Figure \ref{fig:binary} are consistent with the results for continuous outcomes. In Setting 2 with p = 1000, TSGBT performed close to FR regarding sCORR. Overall TSGBT performed better than all other approaches in both sCORR and MSE, as shown from Figures \ref{fig:binary1scorr}, \ref{fig:binary2scorr}, \ref{fig:binary1mse} and \ref{fig:binary2mse}. For example, the median values for sCORR under Setting 1 were 0.67(0.64), 0.66(0.63), 0.66(0.63), 0.74(0.61), 0.75(0.68), 0.71(0.49)  for FR, MC, MC-aug, SGBT, TSGBT and GBT under p = 50(1000), respectively; the median values for MSE under Setting 2 were 3.03(3.41), 2.93(3.33), 3.12(3.92), 3.59(4.08), 2.20(2.77), 2.34(3.41) for FR, MC, MC-aug, SGBT and TSGBT under p = 50(1000), respectively. In Figure \ref{fig:binary3mse}, TSGBT nearly outperforms all other methods in MSE under the null interaction setting, except that FR performs similar to TSGBT when p = 50. We next explored more null scenarios with non-zero main effects and zero interaction signals in Section \ref{permutationtest}.

\subsection{Permutation test}
\label{permutationtest}
We conducted simulations to evaluate the performance of the proposed permutation test in controlling the false positive rate. For a simulated dataset, we fixed the estimated effects from the first stage and implemented the second stage on the permuted covariates as proposed in Section \ref{permutation-metho}. We repeated the permutation 1000 times to obtain a permutation distribution for the variability of the heterogeneous treatment effects of the training samples and calculated p-values, the proportions of permutations with the variability greater than or equal to the observed variability. To evaluate the type I error, we generated 1000 simulated datasets from the model without interaction effect.

We generated N=300 from the model,
$Y=(\alpha_0+\sum_{j=1}^p\alpha_jX_j)^2+\beta_0\cdot T+\sigma_0\epsilon,$
where $\beta_0=0.8$, $\sigma_0=2$ and $\epsilon \sim \mathcal{N}(0,1)$, and generated N=1500 from the primary exponential model for relative risk, $\log{\frac{P_1}{P_{-1}}}=\beta_0$ with its nuisance model for the odds product, $\log{\frac{P_1P_{-1}}{(1-P_1)(1-P_{-1})}}=-C-(\alpha_0+\sum_{j=1}^p\alpha_jX_j)^2$ \citep{richardson2017modeling}, where both $P_1=\mathbb{E}[Y|X,T=1]$ and $P_{-1}=\mathbb{E}[Y|X,T=-1]$ are derived from the specified primary and nuisance models
, $C$ is chosen similarly as in Section \ref{binaryoutcome-est}, and $\beta_0=0.3$. We varied the parameters for the main effects, while keeping the other parameters same. We consider three scenarios with strong, weak and no main effects: In {\bf P1}, $\alpha_0=(\sqrt{3})^{-1}, \alpha_j=(2\sqrt{3})^{-1}, j=3,4,\dots, 10$ with other $\alpha$'s being zero, $\rho=0.5$; In {\bf P2}, $\alpha_0=(\sqrt{6})^{-1}, \alpha_j=(2\sqrt{6})^{-1}, j=3,4,\dots, 10$ with other $\alpha$'s being zero, $\rho=0.5$;
In {\bf P3}, $\boldsymbol{\alpha}=(0,0,\dots, 0)$, $\rho=0.5$. As shown from Table \ref{tab:permutationp501000}, the proposed permutation test for TSGBT protects the correct level of the false positive rate in all the scenarios with strong, weak, no main effects.

\begin{table}
\centering
\caption{Type I error of the permutation test under $p=50$ and $p=1000$}
\begin{tabular}{ccccccccc}
\hline\hline
&Significance&\multicolumn{3}{c}{p=50}&&\multicolumn{3}{c}{p=1000}\\
\cline{3-5}\cline{7-9}
&Level&P1&P2&P3&&P1&P2&P3\\
\hline
Continuous&0.01&0.007&0.007&0.011&&0.011&0.010&0.013\\
&0.05&0.051&0.051&0.055&&0.044&0.055&0.053\\
&0.10&0.100&0.102&0.104&&0.094&0.112&0.108\\
\hline
Binary&0.01&0.009&0.009&0.007&&0.008&0.012&0.011\\
      &0.05&0.047&0.054&0.046&&0.041&0.049&0.053\\
      &0.10&0.098&0.107&0.087&&0.083&0.098&0.103\\
\hline
\end{tabular}
\label{tab:permutationp501000}
\end{table}

\section{PCPT Data Analysis}
\label{sec:PCCT}

Prostate Cancer Prevention Trial (PCPT) was conducted by the SWOG to test whether finasteride can prevent prostate cancer (PCa) (ClinicalTrials.gov NCT00288106). Patients were randomly assigned into the placebo or the finasteride groups with equal probability \citep{thompson2003influence}. The primary endpoint was the biopsy-detected PCa in the 7-year period, where these men were defined as cases. The controls were defined to be those men who completed the prostate biopsy with no evidence of PCa at the end of the study. A case-control study was conducted for studying genetic factors and risk of PCa, and controls were sampled with frequency matched to cases on distributions of treatment arm, age, and positive family history for a first-degree relative with PCa \citep{goodman2010transition}. 

In our analysis, we focused on the risk of the high-grade PCa, the clinically meaningful endpoint, and retrieved genotypic data from 306 high grade cases and 1,365 controls who had sufficient DNA from white blood cells available for genotyping in various projects, which have been
published previously with details in genotyping \citep{price2016association,chen2016adding,chu2018circadian}. Genotypic data contained a total of 444 SNPs from 85 candidate genes, among which 44 SNPs were filtered out by the quality-control metric for the Hardy-Weinberg test. The heterogeneous treatment effect of finasteride is defined as the risk ratio of developing high-grade PCa in a genotype subgroup. We aim to develop a predictive signature for personalized treatment effects using these genotypes. We applied the two-stage gradient boosting trees approach to the data with case-control sampling weights. The sampling weights for controls are $W_i=7,110/1,365\approx 5.21$ and for cases are $W_i=1$. We included the inverse sampling probability $W_i$ in both first-order and second-order gradient statistics, as defined in $G^{a}, H^{a}$ and $G, H$ in Section \ref{tsgbt-metho}.

\begin{figure}
\centering
\subfigure[Diagnostic plot for main-effect]{%
\includegraphics[width=0.45\linewidth, height=5cm]{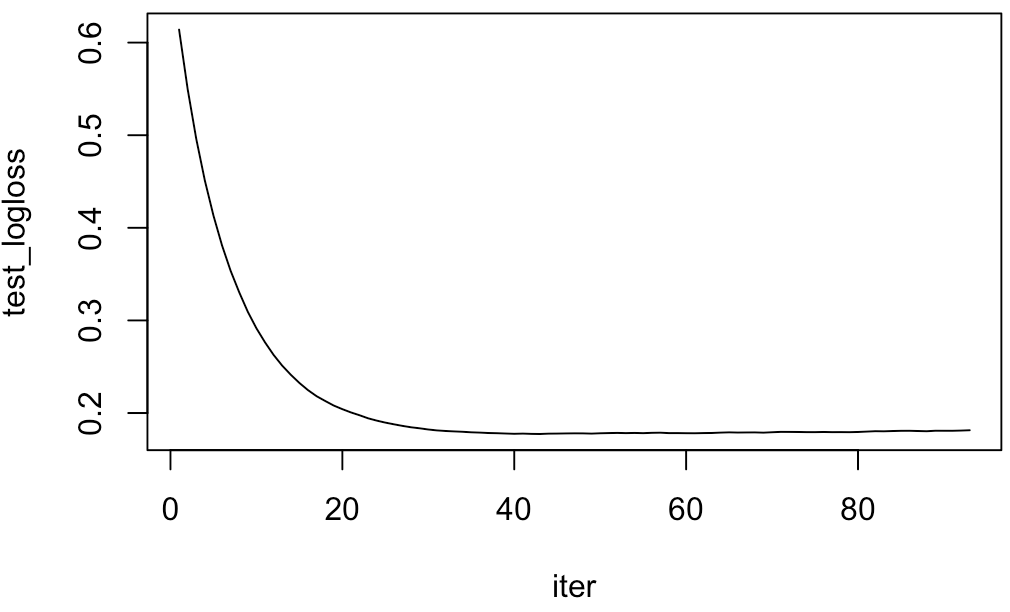}
\label{fig:main-diag}}
\quad
\subfigure[Diagnostic plot for HTE]{%
\includegraphics[width=0.45\linewidth, height=5cm]{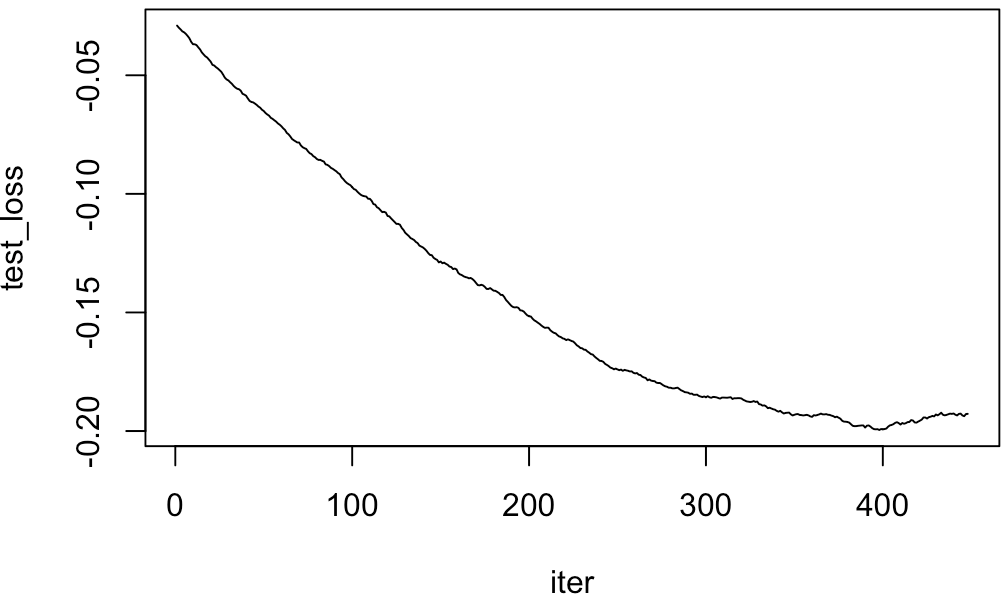}
\label{fig:HTE-diag}}
\subfigure[Individualized treatment effect]{%
\includegraphics[width=0.4\linewidth, height=5cm]{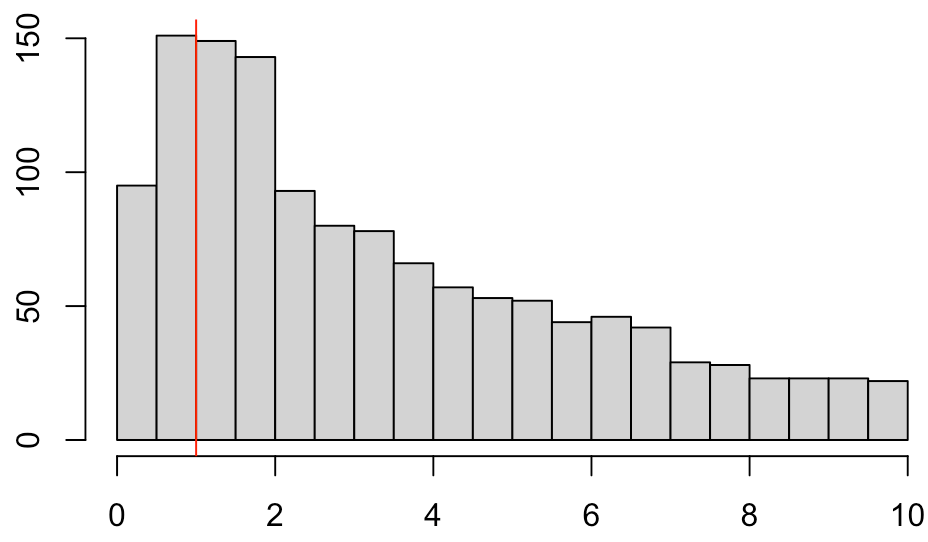}
\label{fig:ITE}}
\quad
\subfigure[Top 20 variables by variable importance]{%
\includegraphics[width=0.4\linewidth, height=5cm]{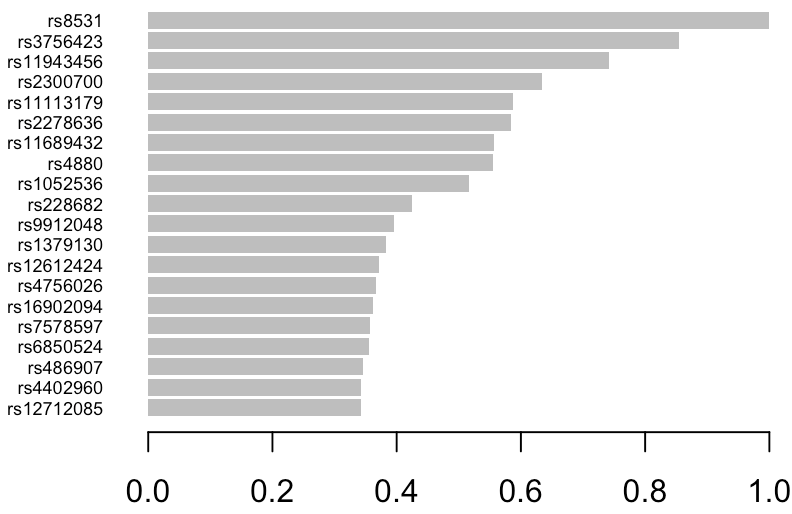}
\label{fig:VI}}
\subfigure[Variable importance gains]{%
\includegraphics[width=0.45\linewidth, height=5cm]{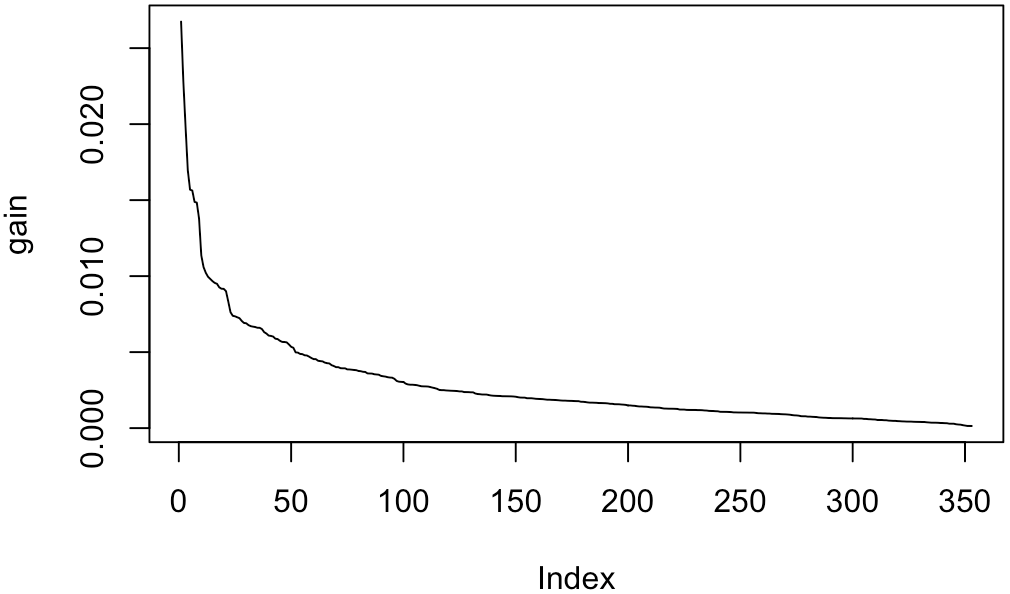}
\label{fig:VI_gain}}

\caption{Diagnostic plots between the number of iterations and the testing log-loss/loss in XGBoost models of the main-effects/HTE based on those 400 SNPs on the risk of developing high-grade prostate cancer, histogram of risk-ratio HTEs based on those 400 SNPs for these 306 high-grade cases and 1,365 controls, relative variable importance plot of top 20 SNPs in contributing to estimate HTE, and variable importance loss gains plot for all 400 SNPs.}
\label{fig:PCPT_ITE_VI}
\end{figure}

In the first stage, we chose the hyperparameter values from the commonly used ranges, where \emph{max-depth} = 6, $\eta=0.1$, $\gamma=4$, \emph{colsample} = 0.7, \emph{subsample} = 0.6, \emph{min-child-weight} = 2. The number of iterations $M_a$ was tuned by ten-fold cross-validation with early-stopping to be 43 and its diagnostic plot presented in Figure \ref{fig:main-diag} suggests weak main effects of SNPs on PCa. In the second stage, we tuned these hyper-parameters by sequential ten-fold cross-validation, which were $\eta=0.01$, \emph{max-depth} = 4, \emph{colsample} = 0.7, \emph{subsample} = 0.6, and selected large values for $\gamma=8$, \emph{min-child-weight} = 12 to be conservative. We then tuned $M$ via ten-fold cross-validation with early-stopping to be 391. The diagnostic plot in Figure \ref{fig:HTE-diag} suggests that these SNPs may interact with finasteride to influence the risk of high-grade PCa. Figure \ref{fig:ITE} shows that the proposed TSGBT method classified 14.7\% of men who have reduced risk of high-grade PCa when taking finasteride. We presented the top 20 genetic variants ranked by the relative variable importance in Figure \ref{fig:VI}. The raw variance importance gain plot for all selected 350 SNPs is displayed in Figure \ref{fig:VI_gain} from the largest to the smallest. The derivation of variable importance in {\tt XGBoost} is shown in Supplement S3. The gains decay steeply with the first 100 SNPs and subsequent SNPs make very minimal contribution.

Potential overfitting might occur in using the entire data to perform estimation. To alleviate  this potential problem, we partitioned data into 90\% training and 10\% testing sets, applied the trained model to the testing data to estimate the proportion. We repeated this cross-validation procedure multiple times and observed that the proportions of patients with a reduced risk of developing high-grade PCa when taking finasteride were consistent to the estimation using the whole dataset.

We implemented the proposed permutation test of no heterogeneity in individualized treatment effects. We used the median absolute deviation (MAD) to  measure the robust variance of the estimated individualized treatment effects in the risk-ratio scale. The p-value from the permutation test was 0.051, suggesting moderately significant heterogeneity of the treatment effects of finasteride in preventing the risk of high-grade prostate cancer.

\section{Discussion}
\label{sec:disc}
For both continuous and binary trial end points, we have proposed a flexible two-stage statistical learning procedure for estimating HTE while gaining efficiency from augmentation. This procedure allows optimal augmentation terms derived for an arbitrary interaction model under consistent nonparametric learning for HTE. Thus, a simple and efficient two-stage gradient boosting trees algorithm is proposed for estimating HTE, and is robust to mis-specification of the first-stage model because the two-stage estimators are orthogonal, dictated by structure of randomized treatments. This property makes it possible to develop a permutation test for assessing globally whether there is any evidence of HTE, conditional on the augmentation term.

One natural extension of our work is survival outcome, which is more complicated when there is censoring and event occurring in the follow-up. The simplicity for augmentation term no longer holds, making it a challenging task to implement a two-stage procedure as we constructed for linear and binary outcomes. Another extension is observational study, where we need to replace the randomization probability by propensity scores, and to estimate the propensity score or outcome-weighting to achieve the doubly robust property \citep{nie2021quasi}. Our proposed two-stage boosting procedure can be easily adapted to observational studies under strong ignorability assumptions (i.e. no unmeasured confounders given observed covariates).

\section*{Acknowledgements}
The authors would like to thank Cathy Tangen for providing PCPT genetic data and Lu Tian for helpful comments.\vspace*{-8pt}

\section*{Funding}
This work was supported by the National Institutes of Health grants R01 CA222833. We gratefully acknowledge computing support from National Institute of Health grant S10OD028685.

\section*{Data Availability Statement}
The data that support the findings of this study are available on request from the corresponding author. The data are not publicly available due to privacy or ethical restrictions.

\section*{Supporting information}
Additional supporting information may be found online in the Supporting Information section at the end of this article. The supporting information provides a succinct summary of the modified covariate method (Supplement S1),  the tuning process for the parameters in {\tt XGBoost} (Supplement S2), and expressions for variable importance (Supplement S3),  connections between augmented estimator and maximum likelihood estimator (Supplement S4), parametric simulations showing efficiency and robustness (Supplement S5), as well as extra simulations results under more diverse scenarios (Supplement S6-7).

\appendix

\section{Derivation of the optimal augmentation in Section 2.2}\label{A}
To derive the optimal augmentation term that minimizes the variance of the resulting estimator for $\mathcal{F}(X)$, it is equivalent to derive the augmentation term that minimizes the variance of the augmented estimating function $S^{\text{aug}}(Y, \mathcal{F}(X)T)$, as expressed in Equation (\ref{aug_weighted_conti_est_eq}). In other words,  $a_0(X)$ should minimize
$$\mathbb{E} \left[\Big\{S^{\text{aug}}(Y, \mathcal{F}(X)T)\Big\}^{\bigotimes 2}\right],$$
where by taking its first-order derivative with respect to $a(X)$ and equating to zero, $a_0(X)$ can be obtained via solving the following equation,
\begin{align*}
   \mathbb{E}\Bigg[&\bigg\{S(Y, \mathcal{F}(X)T) - a(X)\bigg(\frac{T+1}{2\text{Pr}(T=1)}+\frac{T-1}{2\text{Pr}(T=-1)}\bigg)\bigg\}\times\\
   &\bigg(\frac{T+1}{2\text{Pr}(T=1)}+\frac{T-1}{2\text{Pr}(T=-1)}\bigg)\bigg|X\Bigg]=0.
\end{align*}
Integrating over the randomization distribution of the treatment, we then obtain the general form for the optimal augmentation term
\begin{align*}
    a_0(X)=&\bigg[\mathbb{E}\Big\{S(Y, \mathcal{F}(X))\Big|X, T=1\Big\} -\mathbb{E}\Big\{S(Y, -\mathcal{F}(X))\Big|X, T=-1\Big\} \bigg]\times\\
&\text{Pr}(T=1)\text{Pr}(T=-1).
\end{align*}
We then apply this general form of the augmentation term for the mean-difference estimand and the risk ratio estimand.

\subsection{The optimal augmentation for mean-difference estimand }\label{A1}
For the mean-difference estimand, we apply the estimating function $S(Y, \mathcal{F}(X)T)$ in Section \ref{continuousresponse},
\[
S(Y, \mathcal{F}(X)T) = \bigg\{\frac{T+1}{2\text{Pr}(T=1)}+\frac{T-1}{2\text{Pr}(T=-1)}\bigg\} \bigg(Y- \mathcal{F}(X) T \bigg)
\]
in the general form of $a_0(X)$ derived in Appendix to obtain the optimal augmentation term under the squared error loss function for mean-difference estimand,
\begin{align*}
    a_0(X) =&  \Big\{\mathbb{E}(Y|X,T=1)\text{Pr}(T=-1)+\mathbb{E}(Y|X,T=-1)\text{Pr}(T=1)\Big\}\\
    & -\mathcal{F}(X)\Big\{\text{Pr}(T=-1)-\text{Pr}(T=1)\Big\}
\end{align*}
As shown in Section \ref{continuousresponse}, $\mathcal{F}(X)$ is defined to be $\frac{1}{2} \Big\{\mathbb{E}(Y|X, T=1)-\mathbb{E}(Y|X, T=-1)\Big\}$. If $\mathcal{F}(X)$ is correctly specified and consistently estimated, the optimal augmentation term becomes $a_0(X) = \frac{1}{2}\Big\{\mathbb{E}(Y|X, T=1)+\mathbb{E}(Y|X, T=-1)\Big\}$. A special case is when $\text{Pr}(T=1)=1/2$,  $a_0(X) =\mathbb{E}(Y|X)$.

\subsection{The optimal augmentation for the risk-ratio estimand }\label{A2}

We apply the estimating function in Section \ref{binaryresponse},
\[
S(Y, \mathcal{F}(X)T) = \bigg\{\frac{T+1}{2\text{Pr}(T=1)}+\frac{T-1}{2\text{Pr}(T=-1)}\bigg\} \bigg\{1-Y - Y\exp\big(-\mathcal{F}(X)T\big)\bigg\}
\]
into the general form $a_0(X)$ derived in Appendix \ref{A} to obtain the optimal augmentation term for the risk-ratio estimand,
\begin{align*}
     a_0(X) = &  1 - \Big[1+\exp\big\{-\mathcal{F}(X)\big\}\Big]\mathbb{E}(Y|X, T=1)\text{Pr}(T=-1) \\
    & - \Big[1+\exp\big\{\mathcal{F}(X)\big\}\Big]\mathbb{E}(Y|X, T=-1)\text{Pr}(T=1)
\end{align*}
As shown in Section \ref{binaryresponse}, $\mathcal{F}(X)$ is defined to be $\log \big\{ \frac{\mathbb{E}(Y|X, T=1)}{\mathbb{E}(Y|X, T=-1)} \big\}$. When $\mathcal{F}(X)$ in Equation (\ref{aug_weighted_bin_est_eq}) can be correctly specified and consistently estimated, the optimal augmentation term becomes $a_0(X) = 1- \mathbb{E}(Y|X, T=1) - \mathbb{E}(Y|X, T=-1)$, where $a_0(X) = 1- 2\mathbb{E}(Y|X)$ when $\text{Pr}(T=1)=1/2$.

\section{The two-stage GBT algorithm in Section 2.2.4}\label{B}
\begin{algorithm}
\SetAlgoLined
$\mathbf{First\ Stage}$\;
 Set initial values $k=1$ and $\mathcal{A}_i^{(0)}$ for $i=1,\dots,n$\;
 \While{$1 \leq k\leq M_a$}{
  $\hat{\text{A}}_{k}=\underset{\text{A}_{k}\in\text{span}(\mathbb{F})}{\text{argmin}}\ \frac{1}{n}\sum_{i=1}^n\bigg\{G^a\big[y_i, \mathcal{A}_i^{(k-1)}\big]\text{A}_{k}(x_i)+\frac{1}{2}H^a\big[y_i, \mathcal{A}_i^{(k-1)}\big]\text{A}^2_{k}(x_i)\bigg\}/\text{Pr}(T=t_i)+\gamma J_{ak}$\;

  $\mathcal{A}_i^{(k)}=\mathcal{A}_i^{(k-1)}+\hat{\text{A}}_{k}(x_i)$ for $i=1,\dots,n$\;
 }
 $\widehat{\mathcal{A}}(x_i)=\mathcal{A}_i^{(M_a)}$, and transform $\widehat{\mathcal{A}}(x_i)$ to obtain $\hat{a}_0(x_i)$ for $i=1,\dots,n$\;
$\mathbf{Second\ Stage}$\;
  Reset initial values $k=1$ and $\mathcal{F}_i^{(0)}$ for $i=1,\dots,n$\;
 \While{$1 \leq k\leq M$}{
  $\hat{\text{F}}_{k}=\underset{\text{F}_{k}\in\text{span}(\mathbb{F})}{\text{argmin}}\ \frac{1}{n}\sum_{i=1}^n \bigg\{G\Big[y_i, \big(\mathcal{F}_i^{(k-1)}\big)t_i\Big]\text{F}_{k}(x_i)+\frac{1}{2}H\Big[y_i, \big(\mathcal{F}_i^{(k-1)}\big)t_i\Big]\text{F}^2_{k}(x_i) -\hat{a}_0(x_i)\text{F}_{k}(x_i)t_i\bigg\}/\text{Pr}(T=t_i)+\gamma J_k$\;
  $\mathcal{F}_i^{(k)}=\mathcal{F}_i^{(k-1)}+\hat{\text{F}}_{k}(x_i)\text{ for }i=1,\dots,n$\;
 }
 Obtain $\widehat{\mathcal{F}}(x_i)=\mathcal{F}_i^{(M)}\text{ for }i=1,\dots,n$ \;
 \KwResult{Heterogeneous treatment effect is the function of $\widehat{\mathcal{F}}(x_i)$ for $i=1,\dots,n$}
 \caption{Two-stage Gradient Boosting Trees}
 \label{tsgbt_algorithm}
\end{algorithm}

\bibliographystyle{unsrtnat}
\bibliography{tsgbt}  






\end{document}